\documentclass[10pt,twocolumn,letterpaper]{article}

\usepackage[pagenumbers]{cvpr} % arXiv

\usepackage{microtype}
\usepackage{cuted}      % strip environment for the two-column teaser (CVPR 2027 kit)
\usepackage{enumitem}
\usepackage{tcolorbox}

\renewcommand{\paragraph}[1]{\vspace{.2em}\noindent\textbf{#1.}}
\usepackage{tikz}
\IfFileExists{plex-sans.sty}{\usepackage{plex-sans}}{\providecommand{\plexsans}{\sffamily}}
\newcommand{\panellabelfont}{\plexsans\fontsize{10.8pt}{13pt}\selectfont\bfseries}
\newcommand{\overlayfig}[2]{%
  \begin{tikzpicture}
    \node[anchor=south west,inner sep=0pt] (img) at (0,0) {#1};
    \begin{scope}[x={(img.south east)},y={(img.north west)}]
      #2
    \end{scope}
  \end{tikzpicture}}

\newcommand{\relabel}[5]{\fill[white] (#1,#2) rectangle (#3,#4);
  \node[anchor=north west,inner sep=1pt] at (#1,#4) {\panellabelfont #5};}

\graphicspath{{figures/}}

\newcommand{\IQm}{I$-$Q\xspace}
\newcommand{\IQp}{I$+$Q\xspace}
\newcommand{\QIQm}{QI$-$Q\xspace}
\newcommand{\QIQp}{QI$+$Q\xspace}

\newcommand{\gemini}{Gemini-2.5-Flash-Lite\xspace}
\newcommand{\gemma}{PaliGemma-3B\xspace}
\newcommand{\llamathree}{LLaMA3.2-11B\xspace}
\newcommand{\qwenseven}{Qwen2.5-7B\xspace}
\newcommand{\qwenthree}{Qwen2.5-3B\xspace}
\newcommand{\gptfourone}{GPT-4.1\xspace}
\newcommand{\gptfourmini}{GPT-4.1-mini\xspace}
\newcommand{\gptfouro}{GPT-4o-mini\xspace}
\newcommand{\llavathirteen}{LLaVA-1.5-13B\xspace}
\newcommand{\llavaseven}{LLaVA-1.5-7B\xspace}
\newcommand{\clipl}{CLIP ViT-L/14-336\xspace}
\newcommand{\dsname}{HumanVQA\xspace}   % dataset name; rename here if needed
\newcommand{\siglip}{SigLIP\xspace}
\newcommand{\coca}{CoCa\xspace}
\newcommand{\clipthirtytwo}{CLIP-B/32\xspace}
\newcommand{\clipsixteen}{CLIP-B/16\xspace}

\newcommand\eatpunct[1]{}

\usepackage{placeins}  % \FloatBarrier, used in the supplement only
\definecolor{cvprblue}{rgb}{0.21,0.49,0.74}
\usepackage[pagebackref,breaklinks,colorlinks,allcolors=cvprblue,linktoc=all]{hyperref}
\usepackage{tocloft}   % room for the A-numbers in the supplement outline
\def\confName{CVPR}
\def\confYear{2027}

\title{\dsname: Human Baselines for VQA Under Controlled Viewing Conditions}

\author{
    Raphi Kang \quad Olivia Wang \quad Katelyn Haly \quad Peiyang Song \\[3pt]
    Hongyi Hu \quad Shinsuke Shimojo \quad Pietro Perona \\[6pt]
    California Institute of Technology \\[2pt]
    {\small\texttt{\{rkang,oywang,khaly\}@caltech.edu}}
}

\begin{document}
% the outline on the supplement's first page lists only the supplement sections
\addtocontents{toc}{\protect\setcounter{tocdepth}{-10}}

%%%%%%%%%%%%%%%%%%%%%%%%%%%%%%%%%%%%%%%%%%%%%%%%%%%%%%%%%%%%%%%%%%%%%%%%%%%%%%%%
%%  PART 1: MAIN PAPER
%%%%%%%%%%%%%%%%%%%%%%%%%%%%%%%%%%%%%%%%%%%%%%%%%%%%%%%%%%%%%%%%%%%%%%%%%%%%%%%%
\makeatletter
\twocolumn[{\@maketitle
%% Teaser figure, typeset inside the title block so that it sits on the first page below
%% the authors; \captionof gives it the figure counter (Fig. 1).
\begin{center}
  \vspace{-0.6em}
  \includegraphics[width=1.0\textwidth]{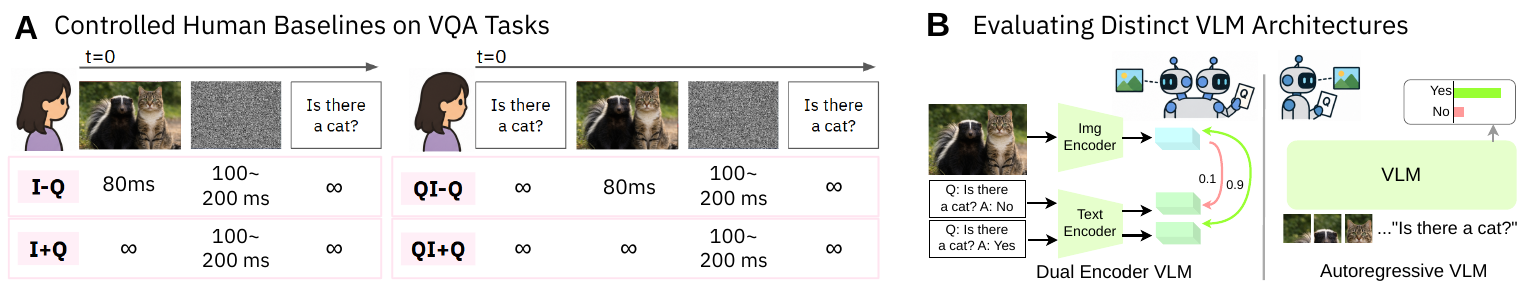}
\end{center}
\vspace{-0.6em}\captionof{figure}{\textbf{Controlled collection of human and VLM responses on the same VQA items.} (A) People answer each VQA sample under four conditions (\cref{tab:terms}). Columns: the question comes only after the image (I, testing gist) or also before it (QI, testing search). Rows: the image is shown for 80~ms ($-$) or until the participant moves on ($+$, $\infty$). Masks follow every image. (B) Two VLM families: dual encoders embed image and question separately, while autoregressive VLMs read them together. Examples in \cref{fig:examples}, human accuracy in \cref{fig:human_acc}.}
\label{fig:overview}
\vspace{1.8em}
}]
\@thanks
\makeatother

%% abstract
\begin{abstract}
% Are vision-language models (VLMs) as accurate as humans in image understanding?
Vision-Language Models (VLMs) are frequently evaluated on benchmarks, which offer human baselines as a guide. How reproducible are these baselines?
We conduct an audit of 101 VQA benchmark reports and find roughly half do not include a human baseline and very few describe the viewing conditions. Further, we show human accuracy can vary greatly based on testing protocol. To address this gap, we introduce \textbf{\dsname}, a dataset of human responses collected under controlled viewing conditions. 153 participants completed 35,136 trials on 960 image-question pairs from ten VQA benchmarks, under four conditions that cross image exposure (80~ms or self-paced) with question order (before or after the image), using a mix of binarized and multiple choice options. In parallel we test 20 VLMs on the same visual questions. We find that viewing conditions change human accuracy by as much as 24 percentage points, and that whether a model exceeds humans depends on human evaluation protocol and model architecture.
Beyond accuracy, we examine whether humans and models tend to err on the same questions. Dual encoder models share error patterns with participants who glimpse the image briefly, about as much as their accuracy alone predicts - while autoregressive VLMs, on average and most clearly among the more accurate ones, share error patterns with participants who view the image at their own pace. We will release the human responses, model outputs and experiment protocol in the hopes of supporting more reproducible and consistent VLM evaluations.
\end{abstract}

%% introduction
\section{Introduction}
\label{sec:intro}

\noindent
Visual Question Answering (VQA)~\cite{agrawal2016vqavisualquestionanswering,goyal_making_2017} and its successors are one of the core capabilities that measure progress in vision-language models (VLMs). Claims that a model is ``near human'' or ``superhuman'' often rest on human baseline comparisons. This score, however, depends largely on the protocol under which it was tested. Research in vision science shows that both viewing time and task knowledge change what people extract from an image. 

A brief glimpse conveys the gist of a scene~\cite{potter1976short,fei-fei_what_2007}, whereas knowing the question and what to look for in advance allows for a targeted search~\cite{treisman1980feature,wolfe_five_2017,yarbus1967eye}. To our knowledge, no previous work
has measured how much this matters for \emph{VLM evaluation}. 
How do evaluation conditions affect human scores on current VQA benchmarks, and does it change the outcome of a model-human comparison? Current reports give us no way to tell. Our audit of 101 VQA benchmark reports (\cref{sec:audit}) found that 52 report no human baseline at all. Of those that did, only one stated a time constraint for its human annotators and only one reported display resolution.

Thus we introduce \dsname, a set of human baselines for VLM evaluation utilizing a controlled viewing protocol (\cref{fig:overview}). Participants answered 960 items from ten VQA benchmarks under four conditions that cross image exposure (80~ms or self-paced) with question order (before or after the image). Our study includes 35,136 trials from 153 participants in a mix of online and in-person settings, with reaction times, workload ratings, and in-person gaze. 
% We also conducted a replication with the original answer options and evaluated 20 VLMs on the same items. 
We will release the experimental materials and protocol, so future researchers can collect comparable baselines.

Just changing the viewing condition has a significant effect. Human accuracy on the same items differed by 14 points overall, and up to 24 points within an individual benchmark (\cref{sec:human}). Viewing conditions also changed which models performed better. 13 of 20 models outscored humans who saw each image for 80~ms, but only 6 outperformed humans who read the question first and viewed the image without a time limit. Published scores, obtained under other protocols, differ even more. For V*, the published human accuracy exceeds every condition we measured. A human score without its protocol therefore cannot support specific claims about task difficulty or model-human comparisons, which motivates a standardized reference protocol (\cref{subsec:protocol}).

A single human baseline is also insufficient for characterizing model errors. Four human references per item let us compare \textit{which} items a model fails on, not only how many (\cref{sec:downstream}). As a demonstration, we find that the item-level errors of dual encoders overlap most with those of participants who glimpsed the image, about as much as their accuracy predicts, while those of autoregressive VLMs, on average, overlap most with participants who viewed the image at their own pace, beyond what accuracy predicts. This difference survives controls for accuracy, answer priors and response format. Further experiments vary input resolution, compare models that share a vision encoder, and test prompting effects. 
Note that human conditions serve mainly as behavioral reference points, and as such we do not draw conclusions about shared mechanisms. Our main contributions are:
\begin{enumerate}[leftmargin=*,nosep,topsep=2pt]
  \item \textbf{A thorough benchmark audit} of human-baseline reporting across 101 VQA benchmark reports (\cref{sec:audit}).
  \item \textbf{The \dsname dataset}: controlled human baselines for 960 items under four documented viewing conditions, with a full-option replication and paired outputs of 20 VLMs, released with the experiment protocol (\cref{sec:dataset}).
  \item \textbf{A measurement of protocol sensitivity} and recommendations for controlled evaluations (\cref{sec:human}).
  \item \textbf{Data analyses comparing VLMs and HumanVQA} demonstrating several human references support item-level analysis of VLM errors (\cref{sec:downstream}).
\end{enumerate}

%% related work
\section{Related Work}
\label{sec:related}

\paragraph{VQA benchmarks}
VQA benchmarks evaluate visual understanding through image-question pairs~\cite{agrawal2016vqavisualquestionanswering,goyal_making_2017}. Recent benchmarks target compositionality (SugarCrepe~\cite{hsieh_sugarcrepe_2023}, ARO~\cite{yuksekgonul_when_2023}, Winoground~\cite{thrush_winoground_2022}), relations (CRPE~\cite{wang2024allseeingprojectv2generalCRPE}), negation (NegBench~\cite{alhamoud_vision-language_2025}), and high-resolution detail (MME-RealWorld~\cite{zhang_mme-realworld_2025}, V*~\cite{wu_v_2023}). Recent works have shown that model scores are sensitive to prompt design~\cite{feng_visually_2026,rosenthal_unexplored_2025} just as human performance depends on task framing~\cite{vriend_which_2024}. We measure the human side of this comparison by using a common item set across viewing conditions.

\paragraph{Humans and models under controlled viewing}
In object recognition, feedforward models account for rapid, masked categorization~\cite{serre2007feedforward,eberhardt2016deep}, while more difficult images recruit recurrent processing~\cite{kar2019evidence}. Viewing time can also provide information on difficulty for humans and models alike~\cite{mayo_how_2023}. Previous work has generally compared model behavior with one human response through error consistency~\cite{geirhos2020beyond,geirhos2021partial,rajalingham2018large}, strategy harmonization~\cite{fel2022harmonizing} and representational alignment~\cite{muttenthaler2023human,sucholutsky2023getting}. \dsname instead provides several references for  each VQA item, making image exposure and question order explicit variables of the comparison.

\paragraph{VLMs and human attention}
Recent work applies controlled behavioral paradigms to VLMs~\cite{tangtartharakul2026visual,zhang2026human} and compares model attention with human attention in VQA~\cite{das_human_2016,sood_vqa-mhug_2021} or with human gaze~\cite{yamamoto_emergence_2024}. These comparisons often find model attributions to be less precise and less task-driven~\cite{bonnen_evaluating_2024,van_dyck_comparing_2021}. Dual encoders are also known to discard relational detail~\cite{yuksekgonul_when_2023,kang_is_2025}. We build on this work by asking whether different VLM architectures produce errors similar to those of participants tested under the same viewing conditions.

%% audit
\section{How Are Human Baselines Reported Today?}
\label{sec:audit}

\begin{figure}[t]
  \centering
  \includegraphics[width=\linewidth]{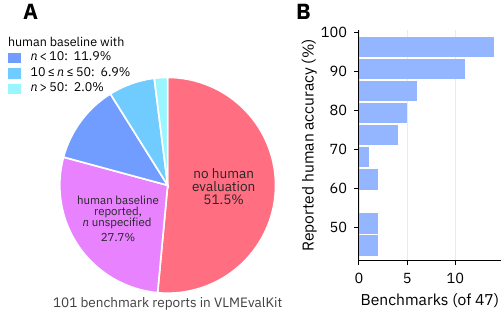}
  \caption{\textbf{Human baselines are often missing or under-reported.} (A) Reporting of human baselines in the 101 benchmark reports in VLMEvalKit. (B) Reported accuracies for the 47 that state one.}
  \vspace{-8pt}
  \label{fig:audit}
\end{figure}

We reviewed 101 papers that introduce the VQA benchmarks in VLMEvalKit~\cite{duan_vlmevalkit_2025} as of January 2026. For each paper, we recorded whether a human baseline exists, the number of human participants, accuracy, and the viewing details including screen size, question order, time limits, viewing distance and image resolution. An LLM helped locate relevant sections and every entry was manually checked (\cref{supp:audit}).

\begin{figure*}[t]
  \centering
  \includegraphics[width=\textwidth]{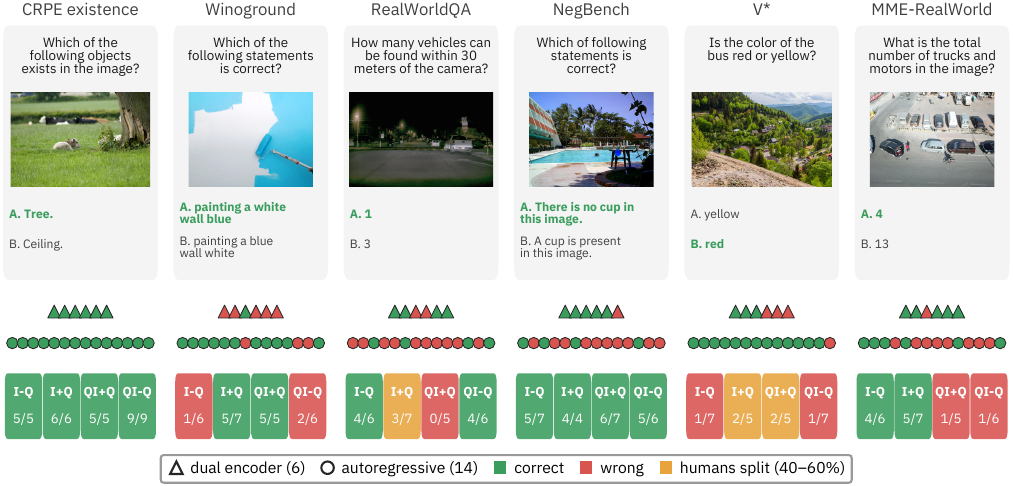}
  \caption{\textbf{Example items and responses.} Six items from different benchmarks (correct option in green). Below each item: the answers of the 6 dual encoders (triangles) and the 14 autoregressive VLMs (circles), and the human consensus in each viewing condition with the number of participants who answered correctly. The same item can be easy after a glimpse and hard under guided search (MME-RealWorld example), or the reverse (Winoground example), and model families often fail on different items.}
  \label{fig:examples}
\end{figure*}

Half of the reports (51.5\%) include no human evaluation, and another 27.7\% report a human score without specifying the number of participants (\cref{fig:audit}A). Viewing details are almost never given: just one report states a time constraint for its annotators (MMLongBench-Doc~\cite{ma2024mmlongbenchdoc}) and one gives the screen resolution (V*~\cite{wu_v_2023}), although both strongly affect human visual performance~\cite{potter2014detecting,mayo_how_2023}. The reported scores are mostly high (40 of 47 exceed 70\%; \cref{fig:audit}B) and often motivate the benchmark itself. This is of course a gap in documentation, not evidence that these human studies were poorly controlled. Their protocols may have been careful, but cannot be recovered from the reports.

%% dataset
\section{The \dsname Benchmark}
\label{sec:dataset}

\dsname tests each item under all four viewing conditions, calibrates the physical size of the displayed image for each participant, and pairs human responses with model outputs for item-level comparison. 

\subsection{Experiment Setup}

\paragraph{Stimuli}
We randomly sampled 96 items from each of ten benchmark subsets: MMVP~\cite{tong_eyes_2024}, MME-RealWorld~\cite{zhang_mme-realworld_2025}, WhatsUp~\cite{kamath_whats_2023}, NegBench~\cite{alhamoud_vision-language_2025}, SugarCrepe~\cite{hsieh_sugarcrepe_2023}, Winoground~\cite{thrush_winoground_2022}, RealWorldQA~\cite{realworldqa2024}, V*~\cite{wu_v_2023}, and the existence and relation subsets of CRPE~\cite{wang2024allseeingprojectv2generalCRPE}. These benchmarks offer two to five options, so raw accuracies would sit on different chance levels (50\% to 20\%) and could not be compared or pooled across benchmarks. We therefore reduce each item to a two-alternative choice, keeping the correct answer and one randomly chosen incorrect option. For the four benchmarks with more options, we conduct a separate study extension which restores the full set of options (\cref{subsec:format}). \Cref{fig:examples} shows six example items, together with the corresponding human and model responses.

\paragraph{Viewing conditions}
\label{subsec:conditions}
\begin{table}[h]
  \centering
  \footnotesize
  \setlength{\tabcolsep}{4pt}
  \begin{tabular}{@{}llll@{}}
    \toprule
    Condition & Image shown & Question shown & Plain name \\
    \midrule
    \IQm & 80 ms & after image & glimpse \\
    \QIQm & 80 ms & before image & cued glimpse \\
    \IQp & self-paced & after image & free viewing \\
    \QIQp & self-paced & before image & guided search \\
    \bottomrule
  \end{tabular}
  \caption{\textbf{The four viewing conditions.} A pattern mask follows every image; the question is shown again after the mask.}
  \vspace{-4pt}
  \label{tab:terms}
\end{table}
Each trial varied two factors (\cref{tab:terms}). First, image exposure where the image is presented for 80~ms, enough for the gist of a scene to be captured but generally too brief for eye movements~\cite{robinson1964mechanics,potter2014detecting}, or displayed until the participant chose to continue (self-paced). Second, question order where the question appears after the image (I) or before it (QI). A 100 or 200~ms pattern mask follows every image, and images are shown at three physical sizes (2.5, 4.2 and 6.7~inches wide at approximately arm's length).

\begin{figure*}[t]
  \centering
  \includegraphics[width=0.98\textwidth]{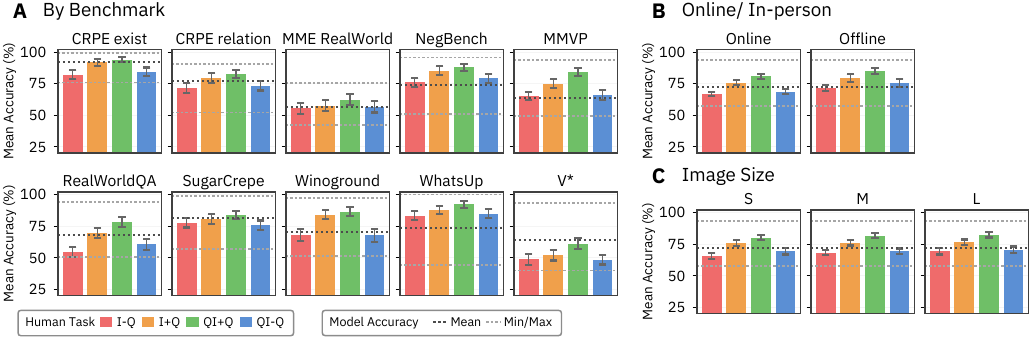}
  \caption{\textbf{Human accuracy depends on viewing condition and benchmark.} Mean human accuracy in each of the four conditions (A) for each of 10  benchmarks, (B) online and in person (``Offline''), and (C) at each  of three image sizes. Error bars indicate 95\% CIs. Dotted lines mark the mean, minimum and maximum accuracy of the 20 models. Guided search (\QIQp, green) is the easiest condition in every panel.}
  \label{fig:human_acc}
\end{figure*}

\subsection{Participants and procedure}
\label{subsec:procedure}
\paragraph{Online study} 100 participants recruited through Prolific completed the experiment in Gorilla~\cite{anwyl2020gorilla}. Participants first matched an on-screen rectangle box to a credit card, allowing us to keep the physical image size fixed and consistent across monitors ~\cite{mayo_how_2023}. Participants then answered 240 items in eight blocks of 30 (one per combination of condition and mask), rating workload on the NASA Task Load Index~\cite{hart1988development} after each block. We used a counterbalanced design of 96 session versions so that each item appeared in every condition.

\paragraph{In-person study} 20 further participants completed the same experiment on a calibrated monitor with a chin rest, while a webcam recorded gaze with WebGazer.js~\cite{papoutsaki2016webgazer} at 30~Hz.

\paragraph{Full-option study} To test whether the two-option format drives our results, we recruited another 34 online Prolific participants and excluded one for clicking through, leaving 33. They answered 192 items with the source benchmarks' original two to five answer options under all four viewing conditions. Four of our benchmarks (MME-RealWorld, RealWorldQA, V* and WhatsUp) have more than two original options, and we took half of each one's 96 items (48 each).

\paragraph{Quality control} Participants who could not complete calibration or the experimental session in time were rejected. Included participants failed at most two of eight attention checks, and 84 of 100 passed all eight. Because brief presentations are hard to control in a browser, we checked the logged durations. The 80~ms image lasted a median of 83.9~ms online (5th-95th percentile 80.3-85.1~ms) and 87.4~ms in person (69-121~ms). Every item-condition cell has 4-9 online responses (median 6). Recruitment, devices and timing details appear in \cref{supp:protocol}. All studies followed an IRB-approved protocol.

\subsection{Model evaluation}
\label{subsec:models}
We evaluated 20 VLMs from two architectural families. \emph{Autoregressive VLMs} feed image tokens to a language model together with the question: \llamathree~\cite{grattafiori_llama3_2024}, \llavaseven and \llavathirteen~\cite{liu2024improved}, \gemma~\cite{beyer2024paligemma}, \qwenthree and \qwenseven~\cite{bai_qwen25vl_2025}, InternVL2.5-1B~\cite{chen2024expanding}, nanoLLaVA~\cite{nanollava2024}, \gemini~\cite{comanici_gemini_2025}, Gemini-3.8-Flash~\cite{google2026gemini38}, \gptfouro, \gptfourmini, \gptfourone~\cite{openai_gpt-4_2024} and gpt-6-luna~\cite{openai2026gpt6}. \emph{Dual encoders} embed images and text separately and pick the highest scoring pair: \clipsixteen, \clipthirtytwo~\cite{radford_learning_2021}, \siglip~\cite{zhai_sigmoid_2023}, \coca~\cite{yu_coca_2022} and BLIP~\cite{li2022blip}, scored by its contrastive head (BLIP-ITC) or its image-text matching head (BLIP-ITM, which attends from text to image but has no language model). Correctness is averaged over up to five prompt templates per model (\cref{supp:models}).

\paragraph{Data release}
\label{subsec:release}
We will release paths to all 960 stimuli with both original and two-option answer sets; all 28,800 two-way and 6,336 full-option trials with viewing condition, response, correctness and reaction time; participant workload ratings and gaze recordings; model outputs from every evaluation; the audit table; and the Gorilla experiment and analysis code. A datasheet and licenses are in \cref{sec:supp_datasheet}.

%% human results
\section{Human Baselines by Protocol}
\label{sec:human}

\paragraph{How much does the viewing condition matter?\eatpunct}
\label{subsec:human_sensitivity}
\begin{table}[t]
  \centering
  \footnotesize
  \setlength{\tabcolsep}{4pt}
  \begin{tabular}{@{}lcccccc@{}}
    \toprule
    & \multicolumn{4}{c}{All benchmarks, binary} & \multicolumn{2}{c}{Multi-choice} \\
    \cmidrule(lr){2-5}\cmidrule(l){6-7}
    Condition & Acc. & Agree & Unan. & $\kappa_{\mathrm{LOO}}$ & two & orig. \\
    \midrule
    \IQm (glimpse) & 66.9 & 75.0 & 21.6 & 0.37 & 59.2 & 41.1 \\
    \QIQm (cued glimpse) & 68.7 & 80.0 & 24.2 & 0.39 & 62.3 & 41.5 \\
    \IQp (free viewing) & 75.8 & 83.3 & 34.6 & 0.51 & 67.2 & 51.0 \\
    \QIQp (guided search) & 81.0 & 85.7 & 44.8 & 0.63 & 73.6 & 57.8 \\
    \bottomrule
  \end{tabular}
  \caption{\textbf{\dsname human baselines.} Accuracy (\%) over all online trials (intervals in \cref{tab:supp_contrasts}). Agree: median share of participants giving the modal answer to an item. Unan.: share of item-condition cells with unanimous answers. $\kappa_{\mathrm{LOO}}$: mean agreement of one participant with the majority of the others. Last two columns: accuracy on the 192 full-option items with two options and with the original options (chance 50\% and about 25\%).}
  \label{tab:baselines}
\end{table}
\Cref{tab:baselines} summarizes human performance across the viewing conditions and \cref{fig:human_acc} shows results by benchmark, setting and image size. Viewing time has the larger effect: self-paced viewing increased accuracy by 8.8 points relative to an 80~ms glimpse. Showing the question in advance adds a further 5.2 points during self-paced viewing, but only 1.8 points when exposure is limited to 80 ms. Knowing what to look for thus helped most when participants had time to look.
% TODO (interaction test): report the exposure x order interaction from the logit model, e.g.
% ``The exposure $\times$ order interaction is X points (95\% CI A-B).'' If the CI includes zero, keep the softened sentence above.

Across all items, the accuracy gap between the lowest and highest-performing conditions is 14.0 points overall (95\% CI 12.5-15.6), with a median of 13.3 and a maximum of 24.4 points (RealWorldQA) across the panels of \cref{fig:human_acc}. Guided search is the highest-performing condition on every benchmark. The effect is larger for single items. For example, the Winoground item in \cref{fig:examples} is solved by all 5 people under guided search but by only 1 of 6 after a glimpse. Participants also agreed most strongly with each other under guided search (\cref{tab:baselines}).
Image size had a smaller effect, where accuracy varied by only 2.7 points between images shown from 2.5 to 6.7~inches wide, and we found no effect of mask duration. We provide further details in \cref{supp:human_results}.

\paragraph{Implications for human-model comparison}
In the glimpse condition, 13 of our 20 models score higher than humans, but in the guided search condition, only 6 outperform humans (Gemini-3.8-Flash, GPT-6-Luna, \qwenseven, \gptfourone, \gptfourmini, \qwenthree). Therefore, the same models on the same items can appear above or below human performance depending on a viewing protocol detail that most benchmark reports do not state.

\paragraph{Answer format robustness}
\label{subsec:format}
\begin{figure}[t]
  \centering
  \includegraphics[width=\linewidth]{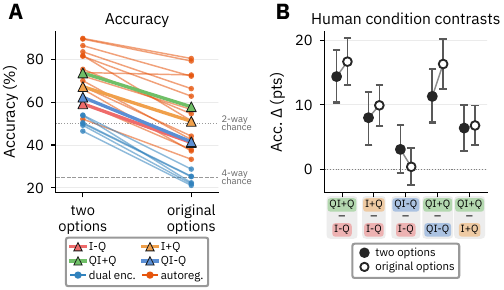}
  \caption{\textbf{Condition effects survive a change of answer format.} On 192 items answered with two options and with the benchmarks' original 2-5 options: (A) accuracy of the four human conditions (triangles) and of each model (dots); (B) differences between human conditions, with 95\% bootstrap intervals.}
  \label{fig:format}
\end{figure}
On the 192 full-option items (\cref{fig:format}), restoring the original options lowers human accuracy by 16-21 points and model accuracy by 18-26, with dual encoders falling to chance. The condition effects are unchanged: guided search exceeds a glimpse by 14.4 points with two options and 16.7 with the original options, and only one of five contrasts changes reliably (\cref{supp:fullopt}). This supports the robustness of the viewing-condition effects.

\paragraph{Comparison with published human baselines}
\label{subsec:published}
\begin{table}[h]
  \centering
  \footnotesize
  \setlength{\tabcolsep}{4pt}
  \begin{tabular}{@{}lccccc@{}}
    \toprule
    & & \multicolumn{4}{c}{\dsname (96 items, two options)} \\
    \cmidrule(l){3-6}
    Benchmark & Published & \IQm & \QIQm & \IQp & \QIQp \\
    \midrule
    MMVP~\cite{tong_eyes_2024} & 95.7 & 65.1 & 67.4 & 75.2 & 82.5 \\
    Winoground~\cite{thrush_winoground_2022} & 89.5 & 66.5 & 70.0 & 81.7 & 87.5 \\
    V*~\cite{wu_v_2023} & 98.95 & 48.0 & 47.8 & 51.7 & 60.6 \\
    \bottomrule
  \end{tabular}
  \caption{\textbf{Published human scores against \dsname.} Published scores are on each benchmark's full item set and original format (Winoground: text score, which matches our caption-choice items; MMVP: accuracy over question pairs, both of which must be correct). \dsname values are mean accuracy (\%) per condition on our 96 items from each benchmark.}
  \label{tab:published}
\end{table}
Three benchmarks in our sample report a single human accuracy score (\cref{tab:published}). These comparisons are approximate, since published scores cover the full benchmark in its original answer format whereas HumanVQA uses 96 sampled items per benchmark. Winoground's published score lies within two points of our guided-search accuracy, whereas MMVP's and V*'s are 13 and 38 points higher, so we could not reproduce them under our protocol. This does not diminish the original work, as these human scores were intended only to show the benchmark tasks were solvable. Our controlled protocol answers a different question, and makes possible comparisons across viewing conditions and item-level analyses that a single reported score cannot support.

\paragraph{Human error signatures}
\label{subsec:human_misses}
\begin{figure}[b]
  \centering
  \includegraphics[width=\linewidth]{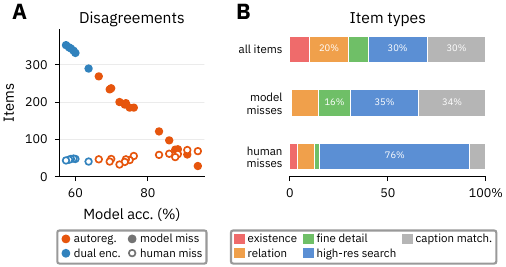}
  \caption{\textbf{Model-Human Disagreements.} (A) Per model, items it misses that the guided-search consensus gets right (filled) and the reverse (open). (B) Item types among all items, model misses and human misses (eleven strong autoregressive models).}
  \label{fig:asym}
\end{figure}
Guided search should not be treated as a ceiling for human performance. Each model answers correctly 32-71 items that the \QIQp human consensus misses (\cref{fig:asym}A). For the eleven strong autoregressive VLMs, 76\% of these items come from V*, MME-RealWorld or RealWorldQA, although these benchmarks only account for 30\% of all items (\cref{fig:asym}B). The V* item in \cref{fig:examples} illustrates this: 13 of 14 autoregressive models answer it correctly, but at most 2 of 5 people do in any condition. Small targets in large images may simply be unresolvable on a fixed-size screen. Model errors show a different pattern, where individual models miss as many as 353 items, with errors spread across many questions. Among the 518 items humans answered correctly in the guided-search condition, only two are missed by nine or more of the eleven strong models (\cref{supp:human_misses}). Thus, we see that human and model accuracy can differ in both magnitude and specific questions.

\subsection{A recommended protocol for VLM evaluation}
\label{subsec:protocol}

\begin{figure*}[t]
  \centering
  \includegraphics[width=0.96\textwidth]{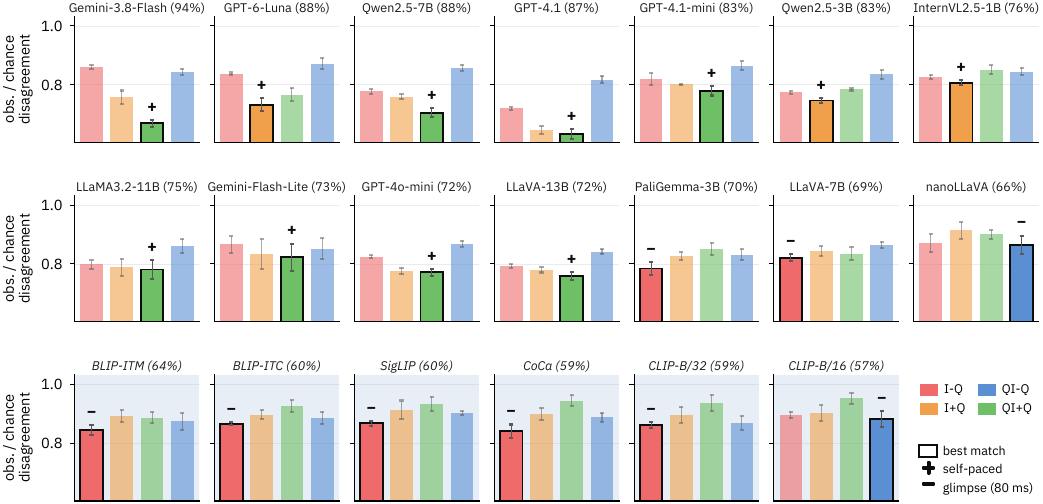}
  \caption{\textbf{Model families' errors overlap most with different viewing conditions.} Disagreement ratio (observed/chance) between each model and each human condition; lower means more shared errors (\cref{subsec:metric}; accuracy in parentheses). Top rows: autoregressive VLMs, bottom row (shaded): dual encoders. Outlined bar: best match ($+$ self-paced, $-$ 80~ms). Error bars: SD across templates.}
  \label{fig:families}
\end{figure*}

We recommend the following when using a human baseline to evaluate VLMs.
\begin{itemize}[leftmargin=*,nosep,topsep=2pt]
  \item \textbf{Use guided search as the standard reference.} Show the question first and let participants view the image at their own pace. Of our four conditions, this best matches how VLMs are evaluated and is the most reliable (\cref{tab:baselines}). It is a standardized reference, not a human ceiling: it compares models with people under a fixed protocol, not with people free to zoom or take unlimited time. Brief-exposure conditions serve as diagnostic references (\cref{sec:downstream}).
  \item \textbf{Match the human and model tasks.} Use the same items, options and scoring; format alone changed human accuracy by 16-21 points.
  \item \textbf{Control what can be seen.} For detail-dependent benchmarks, allow zoom or full-resolution viewing, or label the human score as display-limited.
  \item \textbf{Report the protocol.} Exposure duration, question order, image size and resolution, viewing distance, response format, participants per item, and their agreement.
\end{itemize}

\noindent The released HumanVQA experiment implements these and can be adapted to new benchmark items.

%%  downstream analysis
\section{Using \dsname to Analyze VLMs}
\label{sec:downstream}

As a demonstration of what several human references make possible, we ask \emph{which reference} a model's item-level errors overlap with most. A dual encoder embeds the image before receiving the question, so it might miss the same items as participants who only glimpsed the image; an autoregressive VLM reads image and question together, so it might miss the same items as participants who knew what to look for.
\subsection{Measuring error resemblance}
\label{subsec:metric}
For each item and viewing condition, the human \emph{consensus} is the majority response of online participants (ties count as correct). We count the items on which a model and a consensus disagree and divide by the count expected if the model's errors were random at its accuracy. This \emph{disagreement ratio} approximates $1-\kappa$~\cite{cohen1960coefficient}: 1 is chance, and 0.8 means 20\% fewer disagreements than chance, \ie model and humans tend to miss the same items. The condition with the lowest ratio is the model's \emph{best match}.

For example, with random errors \siglip\ would disagree most with the glimpse consensus (about 430 items) and least with guided search (about 410). Instead it disagrees least with the glimpse consensus (about 370; ratio 0.87; \cref{fig:supp_counts}), which is unlikely unless \siglip\ and glimpse viewers fail on the same items more than accuracy alone predicts (shared item difficulty is addressed in \cref{subsec:not_accuracy}).

The \emph{resemblance index}, ratio(\IQm) minus ratio(\QIQp), is negative when a model's errors overlap more with the glimpse condition and positive when they overlap more with guided search. For our models, $\pm0.05$ corresponds to roughly 5-20 fewer disagreements with one condition than the other, out of 960 items. Uncertainty comes from bootstrapping items (\cref{supp:agreement}).

Agreement is modest. The best model-condition pairs reach $\kappa\approx0.4$, against 0.63 between one participant and the rest under guided search. We therefore ask which condition a model resembles most, not whether it behaves like a human.

\subsection{Two families, two error profiles}
\label{subsec:families}
\Cref{fig:families} shows each model's disagreement ratios. Five of six dual encoders best match the glimpse (\clipsixteen the cued glimpse), about as their accuracy alone predicts (\cref{fig:supp_e19}; \cref{subsec:not_accuracy}). Of fourteen autoregressive VLMs, eight best match guided search, three (GPT-6-Luna, \qwenthree, InternVL2.5-1B) free viewing, and the three least accurate a glimpse condition. The most accurate, Gemini-3.8-Flash, outscores every human condition (93.5\% vs.\ 81.0\% under guided search), yet its errors still overlap most with guided search (bootstrap support 0.97). Whether this family is closer to guided search or free viewing is unresolved - its mean ratios for the two are nearly equal (\cref{fig:notacc}A) and best matches split between them (\cref{tab:supp_boot}), so we describe it as resembling self-paced viewing. Family thus tracks which condition model errors overlap with most, mainly along viewing time. 

Distance between the families' mean profiles is 0.039, against a null mean of 0.004 (95\% range 0.000-0.017) when model labels are permuted and 0.007 (0.000-0.027) when whole lineages are permuted, where it is the most extreme of all 495 labelings ($p=0.002$; \cref{supp:permutation}).

Individual best matches are less certain: the number of autoregressive models best matching guided search ranges from four to eleven across tie rules and agreement statistics (\cref{supp:consensus,supp:reliability}). Every dual encoder keeps its glimpse match under every tie rule, so we draw conclusions at the family level.

\subsection{Is the difference just accuracy?}
\label{subsec:not_accuracy}

All six dual encoder models (57-64\%) are less accurate than all 14 autoregressive models (66-94\%), plausibly because a model that embeds the image before seeing the question is less able to retrieve salient details. A less accurate model's errors might overlap most with the glimpse condition simply by missing more items. We investigate this possibility in three ways, using the \textit{resemblance index} from \cref{subsec:metric}.
% : ratio(\IQm) minus ratio(\QIQp), which is negative when a model's errors look more like a glimpse and positive when they look more like guided search.

\paragraph{A difficulty-preserving null} We shuffled each model's correct responses among items of similar human difficulty. This keeps accuracy and difficulty but removes finer item-level correspondence. Under this null, every VLM has a negative (glimpse-like) index, so accuracy and difficulty alone push models toward the glimpse condition, and the null rather than zero is the right baseline. Nine of the fourteen autoregressive models, including all the strongest, exceed the null's 95\% range, while one dual encoder (\coca) falls below it (\cref{supp:null}). This indicates that the search-like profile of most autoregressive models goes beyond what their accuracy predicts, whereas the glimpse-like profile of most dual encoders is roughly what their accuracy predicts.

\begin{figure}[b]
  \centering
  \overlayfig{\includegraphics[width=\linewidth]{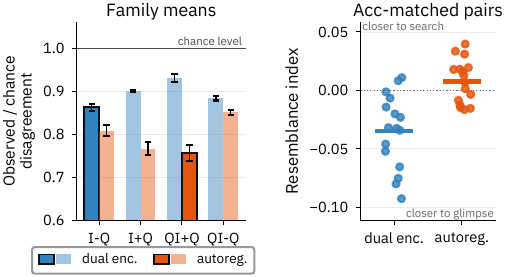}}{%
    \relabel{0.0}{0.93}{0.03}{1.0}{A}
    \relabel{0.50}{0.93}{0.53}{1.0}{B}}
  \caption{\textbf{The family difference is not only an accuracy effect.} (A) Mean disagreement ratio per condition for each family ($\pm$SEM). (B) Resemblance index (ratio(\IQm) $-$ ratio(\QIQp); positive = search-like) for 16 dual-encoder/autoregressive pairs evaluated on identical items with accuracy matched within one point.}
  \label{fig:notacc}
\end{figure}

\paragraph{Matched accuracy} We formed 16 pairs consisting of a dual encoder and a strong autoregressive VLM. For each pair, we removed items that only the autoregressive model answers correctly until the two accuracies agree within one point. On these matched items, no dual encoder best matched guided search, while the autoregressive model kept that match in 6 of 16 pairs and had a higher resemblance index on average (\cref{fig:notacc}B; \cref{supp:matched}). This indicates that the two families fail on different items even when they fail equally often. Because the autoregressive model keeps its guided-search match in only 6 of 16 pairs, part of that match comes with higher accuracy, but the direction of the family difference does not.

\begin{figure}[t]
  \centering
  \includegraphics[width=\linewidth]{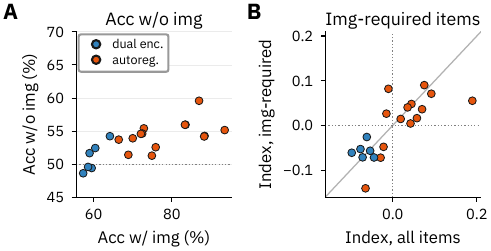}
  \caption{\textbf{The correspondence is not an answer prior.} (A) Accuracy without the image against accuracy with it (dotted: chance). (B) Resemblance index (ratio(\IQm) $-$ ratio(\QIQp); positive = search-like) on all items against the same index on the items a model misses without the image (gray: identity).}
  \label{fig:blind}
\end{figure}

\paragraph{Answer priors} A language model might answer correctly from the question alone. With the image withheld, all models drop to 49-60\% and their errors overlap with no viewing condition in particular (\cref{fig:blind}A). On the items each model fails without the image, the family-level pattern remains unchanged (\cref{fig:blind}B; \cref{supp:blind}). This indicates that the correspondence depends on image information rather than on answer priors in the question text.

\paragraph{Lineage and benchmark} A mixed-effects model with benchmark, model lineage, and accuracy as a covariate agrees (\cref{supp:mixed}), so the family difference is not driven by a few related models or a few benchmarks.

\paragraph{Caveats} Three results temper these controls. In the matched pairs, both models show the expected sign of the index in only 8 of 16 pairs, and the dual encoder's best match splits evenly between \IQm and \IQp (\cref{supp:matched}). In the mixed model, the family effect with accuracy as a covariate is modest ($p=0.048$), and because the families do not overlap in accuracy, its accuracy slope is identified only from spread within each family (\cref{supp:mixed}). Under a reliability-rescaled $\kappa$, only 4 of 14 autoregressive models keep guided search as their best match (\cref{supp:reliability}). Overall, the errors of the strongest autoregressive models are more search-like than their accuracy alone explains, and the glimpse-like behavior of five of the six dual encoders is consistent with what accuracy predicts.

\subsection{Further downstream experiments}
\label{subsec:probes}
We conducted several follow-ups to better understand why the model families differ and what the human-model comparison can be used for. Detailed methods and result breakdowns are included in \cref{supp:probes}.

% \paragraph{Answer format} On the 192 full-option items, the autoregressive family retains guided-search-like profiles when compared against human responses collected in that format (mean index $+0.14$ with two options, $+0.11$ with the originals) and thirteen of fourteen models still best match with the self-paced condition. The dual encoders drop to chance accuracy and no longer show any meaningful profile (\cref{supp:fullopt}). \emph{Takeaway:} the correspondence is not an artifact of binary choice options, but a model must be above chance to have an error pattern profile. Humans and models must be scored in the same answer format.

\paragraph{Image resolution} Reducing \qwenseven from 1{,}280 to 16 visual tokens, or \llamathree from four tiles to one, lowers accuracy (88 to 76\% for \qwenseven) and moves the best match from guided search to free viewing, rather than to a brief glimpse (\cref{supp:resolution}). \emph{Takeaway:} the search-like profile is associated with having enough detail to find question-relevant content. As resolution and accuracy are confounded, we cannot isolate effects. But the error profile offers a diagnostic beyond accuracy, \eg to test whether added resolution is used for question-specific evidence.

\paragraph{Same encoder, different readout} CLIP ViT-L/14 scored contrastively best matches a glimpse condition, whereas \llavathirteen, which reads the same encoder's patch tokens with a language model and the question, best matches guided search. A question-selected crop changes neither a dual encoder's accuracy nor its profile (\cref{supp:resolution}). \emph{Takeaway:} the profile is associated with how the image is read out, not only with the encoder, and cropping before encoding is no substitute for reading the image with the question. As these models also differ in training, scale and preprocessing, only a controlled intervention could establish a causal role.

\paragraph{Instructed emulation} We asked 12 models to predict what a person in a stated viewing condition would answer (\eg after an 80~ms glimpse). The prompts changed 9-60\% of answers but not their error profiles: resemblance indexes moved by less than 0.04, and predictions were no closer to the condition's human responses. Model accuracy varied by at most 3 points across conditions, compared with 14 for people (\cref{supp:emulate}). \emph{Takeaway:} prompting a VLM to simulate a viewing condition therefore does not replace collecting actual human responses.

\paragraph{Per-benchmark breakdown} Dual encoders share the glimpse condition's errors most closely on CRPE, and strong autoregressive models share guided search's errors most closely on CRPE relation, V*, MME-RealWorld and MMVP. Neither family shows alignment on SugarCrepe, NegBench or Winoground (\cref{supp:perbench}). \emph{Takeaway:} model errors overlap with human errors under limited viewing mainly where the task requires finding visual evidence, while compositional and negation errors likely have other causes, such as language understanding (\cf the NegBench item in \cref{fig:examples}). This identifies which benchmarks are informative for such diagnoses.

\paragraph{Gaze} Dual-encoder attribution maps and image-first gaze are both diffuse, while autoregressive attribution and question-first gaze concentrate on task-relevant regions (\cref{supp:gaze}). \emph{Takeaway:} the contrast has plausible spatial signal, though webcam gaze is coarse and preliminary.

%% discussion
\section{Discussion and Limitations}
\label{sec:discussion}

\paragraph{HumanVQA}
Human accuracy depends on protocol as much as on the items. Viewing condition shifts accuracy by up to 24 points and decides whether 7 of our 20 models count as above human. \dsname makes these conditions explicit and provides a reusable protocol.

\paragraph{Downstream VLM experiments}
The two families fail on different items, and the strong autoregressive models' errors overlap with self-paced viewing beyond what accuracy predicts, varying with input resolution and readout but not prompting. These are similarities in which items are missed, not evidence of shared mechanisms. Family also covaries with training objectives, data, scale and preprocessing, and the conditions differ in more than one respect (image-first viewing also requires memory). The dual encoders' glimpse-like profile cannot be separated from their accuracy, so we present it as a description.

\paragraph{Limitations}
The dataset covers 960 2-choice items (192 with full options) with English-fluent adults, not open-ended VQA. Each consensus rests on 4-9 responses, limiting conclusions about single models, and webcam gaze is coarse~\cite{papoutsaki2016webgazer}. Future work includes extension to open-ended questioning and scaling to include a zoom condition, which the human misses call for, and to clarify the V* gap.

%% conclusion
\section{Conclusion}
\label{sec:conclusion}
Human baselines are central to judging VLMs, yet often missing and rarely reproducible. \dsname provides controlled human responses on 960 items from ten benchmarks under four documented viewing conditions, paired with 20 VLMs. Because the protocol alone moves human accuracy by up to 24 points and decides whether a model counts as above human, we recommend a standardized guided-search reference, matched to the model's task and reported with its protocol. Several references per item further turn a single model-human gap into an analysis of which items models fail on. Beyond any particular baseline or architectural finding, we hope \dsname serves as a framework for protocol-aware, multidimensional human evaluation of VLMs.

% %% LLM statement
% \section*{LLM Usage Statement}
% Large language models (Anthropic Claude, OpenAI Codex) were used primarily for coding downstream experiments and manuscript writing assistance. GPT-4o was used for retrieving paper links from huggingface for the benchmark audit, prior to individual human confirmation.
% LLMs were further used for figure editing and internal review passes. LLMs were not used in the idea generation of experiments, or at any stage of the human data collection. 
% All design, data collection, analyses, figures and claims were specified, checked and approved by the authors, who take full responsibility for the content.

\section*{Acknowledgments}
We would like to thank Ilona Demler, Sevan Brodjian, Damiano Marsili, Suzanne Stathatos, Michael Hobley, and Harry Chen for proofreading or helping test pilot experiments. 
Raphi Kang was supported by the NSF Graduate Research Fellowship. Olivia Wang was supported by Caltech SURF and Caltech Associates.

{
    \small
    \bibliographystyle{ieeenat_fullname}
    \bibliography{main}
}

%%%%%%%%%%%%%%%%%%%%%%%%%%%%%%%%%%%%%%%%%%%%%%%%%%%%%%%%%%%%%%%%%%%%%%%%%%%%%%%%
%%  PART 2: SUPPLEMENTARY MATERIAL (single column)
%%%%%%%%%%%%%%%%%%%%%%%%%%%%%%%%%%%%%%%%%%%%%%%%%%%%%%%%%%%%%%%%%%%%%%%%%%%%%%%%
\clearpage
\onecolumn
\setcounter{section}{0}\setcounter{figure}{0}\setcounter{table}{0}\setcounter{footnote}{0}
\renewcommand{\thesection}{A\arabic{section}}
\renewcommand{\thefigure}{A\arabic{figure}}
\renewcommand{\thetable}{A\arabic{table}}
% distinct hyperref anchors for the restarted counters
\renewcommand{\theHsection}{A\arabic{section}}
\renewcommand{\theHfigure}{A\arabic{figure}}
\renewcommand{\theHtable}{A\arabic{table}}
\makeatletter
\begin{center}
  {\Large\textbf{\thetitle}}\\[0.5em]
  {\Large Supplementary Material}\\[1.0em]
\end{center}
\makeatother
% keep the CVPR run-in period out of the supplement outline
\makeatletter
\def\cvprsubsect#1{\cvprsubsection[#1]{\texorpdfstring{\hskip -1em.~}{}#1}}
\makeatother
\noindent\textbf{Organization.} \Cref{supp:audit} documents the reporting audit (main \cref{sec:audit}). \Cref{supp:protocol,supp:human_results} give the full human protocol and additional human results (main \cref{sec:dataset,sec:human}). \Cref{supp:models} describes the model evaluation (main \cref{subsec:models}). \Cref{supp:agreement,supp:controls,supp:probes,supp:gaze} support the downstream analysis of main \cref{sec:downstream}: the agreement metrics, the controls for accuracy, the further probes and the gaze comparison. \Cref{sec:supp_datasheet} is the datasheet. Throughout, \emph{strong} autoregressive models are the eleven with at least 72\% accuracy, and the \emph{resemblance index} is ratio(\IQm) minus ratio(\QIQp), negative for glimpse-like and positive for search-like error profiles.

\vspace{0.5em}
\addtocontents{toc}{\protect\setcounter{tocdepth}{2}}
\setlength{\cftbeforesecskip}{1pt}
\tableofcontents
\clearpage

%% ============================================================================ A1
\section{Audit of Human-Baseline Reporting}
\label{supp:audit}

This section describes the audit summarized in main \cref{fig:audit}.

\paragraph{Scope and fields}
We built a structured table of the reporting practices of the 101 benchmark reports included in VLMEvalKit~\cite{duan_vlmevalkit_2025} as of January 2026. The unit of analysis is a benchmark entry, each linked to a canonical source paper. We recorded 14 fields: dataset name, paper URL, whether a human baseline exists and, if so, the baseline type, participant count, recruitment, setting, whether a score is reported, the context of that score, stimulus presentation, time constraints, screen details, the section containing these details, and free-text notes.

\paragraph{Procedure}
An LLM (GPT-4o) retrieved each report and located candidate sections. An author then verified every field against the source and recorded the supporting section. Fields were marked \texttt{N/A} when no human baseline existed and \texttt{Not Specified} when a baseline existed but the paper omitted the detail. Descriptors are kept as free text so as not to over-normalize heterogeneous reporting.

\paragraph{Summary}
Of the 101 entries, 49 report a human baseline and 52 do not; 47 of the 49 give a numeric score (main \cref{fig:audit}B). Only MMLongBench-Doc~\cite{ma2024mmlongbenchdoc} states a time constraint for its annotators, and only V*~\cite{wu_v_2023} the screen resolution. The table is part of the release.

\FloatBarrier
%% ============================================================================ A2
\section{Human Data Collection}
\label{supp:protocol}

This section expands main \cref{subsec:conditions,subsec:procedure}: the trial structure, calibration, counterbalancing, recruitment and exclusions, display timing, the in-person and full-option studies, and the workload questionnaire.

\subsection{Conditions and trial structure}

\begin{table}[!ht]
\centering
\caption{\textbf{Presentation conditions and stimulus parameters.} Every trial combines one condition with one image size and one mask duration. Question displays and responses are always self-paced.}
\label{tab:supp_notation}
\small
\begin{tabular}{@{}llp{0.42\linewidth}@{}}
\toprule
Shorthand & Name & Trial sequence \\
\midrule
\IQm & Image-first, brief (glimpse) & Image (80~ms) $\rightarrow$ mask $\rightarrow$ question \\
\IQp & Image-first, self-paced (free viewing) & Image (unlimited) $\rightarrow$ mask $\rightarrow$ question \\
\QIQm & Question-first, brief (cued glimpse) & Question $\rightarrow$ image (80~ms) $\rightarrow$ mask $\rightarrow$ question again \\
\QIQp & Question-first, self-paced (guided search) & Question $\rightarrow$ image (unlimited) $\rightarrow$ mask $\rightarrow$ question again \\
\midrule
Parameter & Levels & Definition \\
\midrule
Image exposure & brief ($-$) / self-paced ($+$) & 80~ms / unlimited \\
Mask duration & short / long & 100~ms / 200~ms of visual noise after the image \\
Image size & S / M / L & 2.5 / 4.2 / 6.7 inches wide on a calibrated display at arm's length \\
\bottomrule
\end{tabular}
\end{table}

\Cref{tab:supp_notation} lists the four conditions and the two secondary parameters. A fixation cross preceded every image. Participants advanced self-paced displays with the space bar and answered with the ``A'' and ``B'' keys.

\subsection{Online study}

\begin{figure}[!ht]
  \centering
  \begin{subfigure}{0.32\linewidth}
    \centering
    \includegraphics[width=0.6\textwidth]{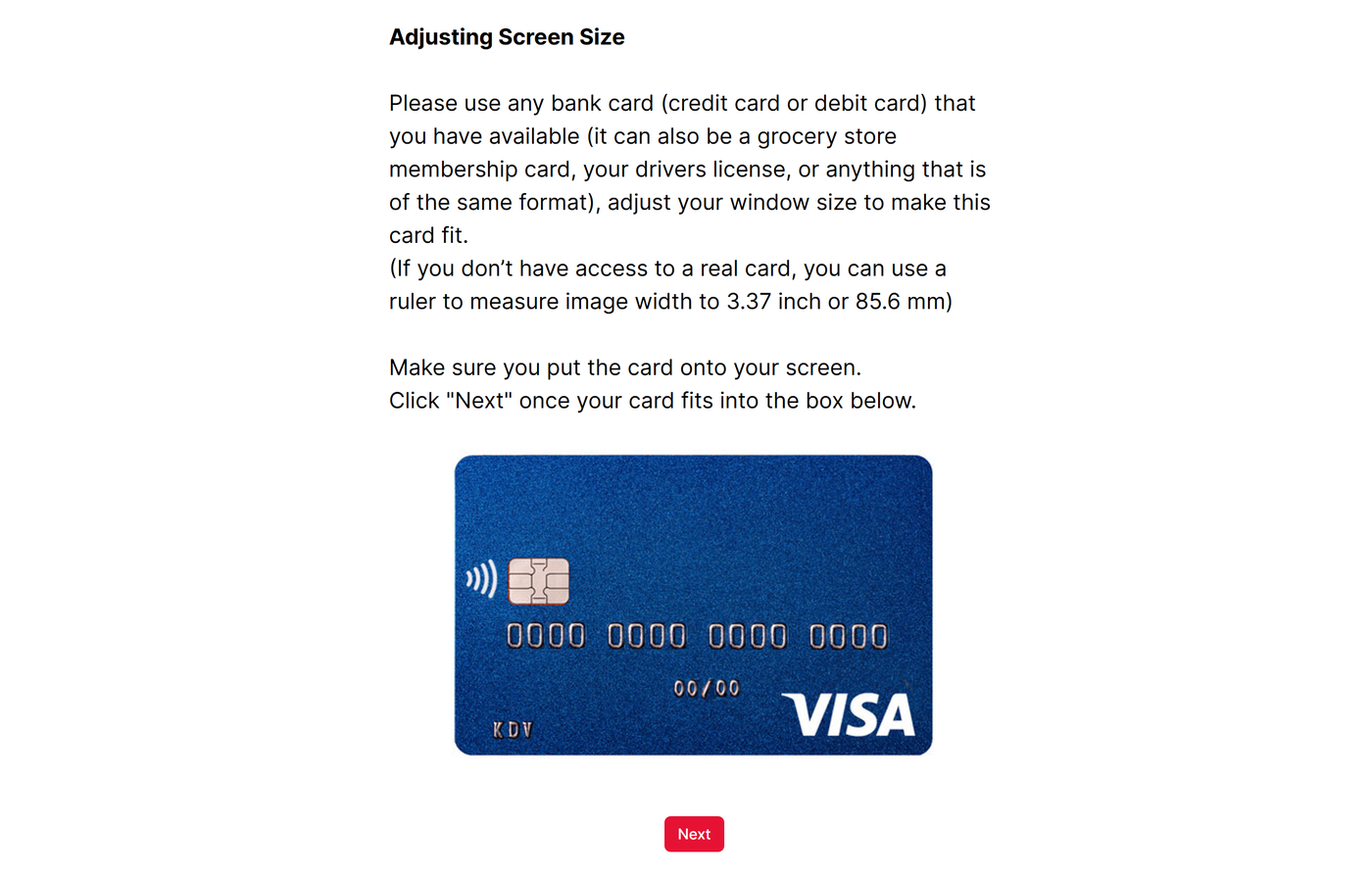}
    \caption{}
  \end{subfigure}\hfill
  \begin{subfigure}{0.32\linewidth}
    \centering
    \includegraphics[width=0.6\textwidth]{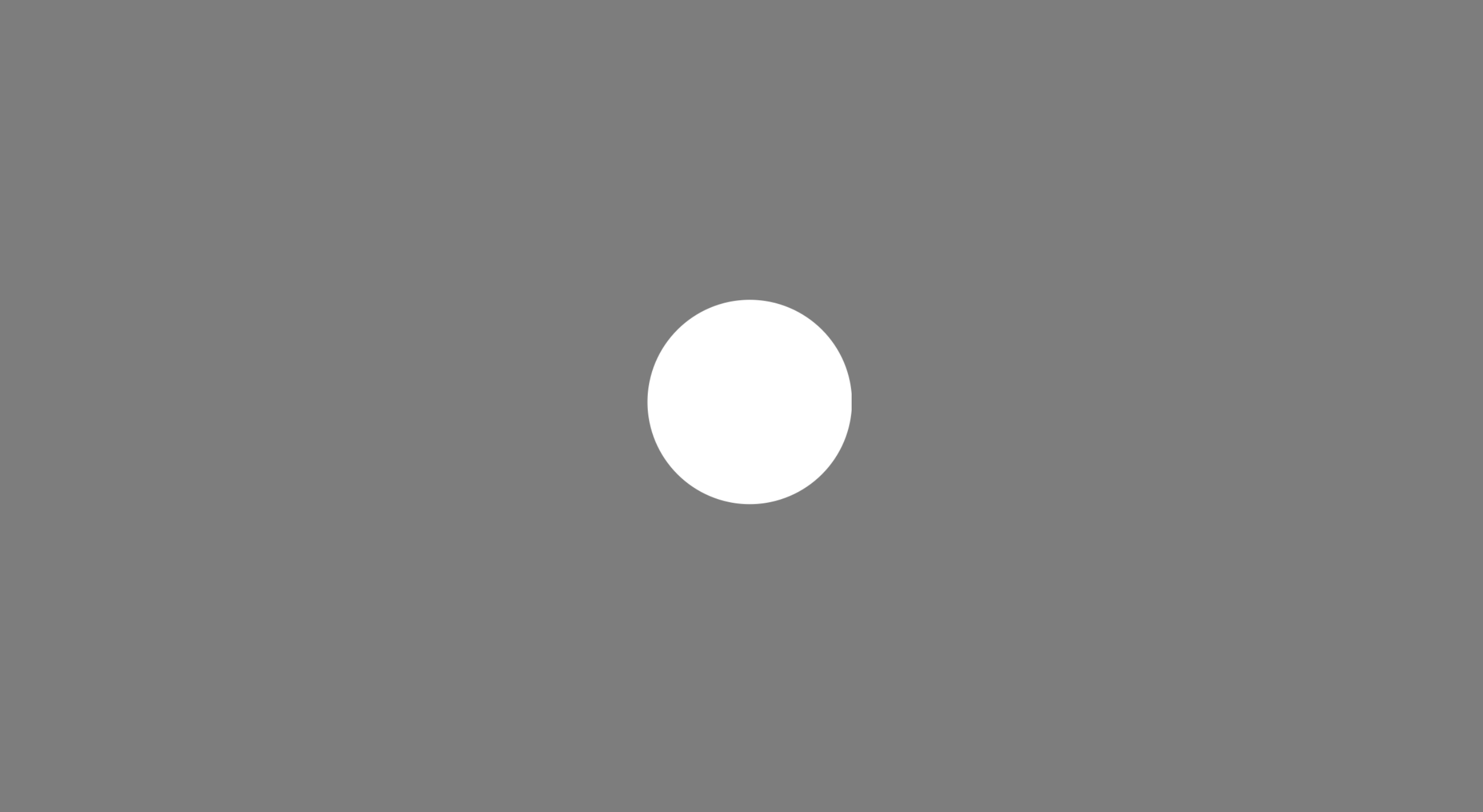}
    \caption{}
  \end{subfigure}\hfill
  \begin{subfigure}{0.32\linewidth}
    \centering
    \includegraphics[width=0.6\textwidth]{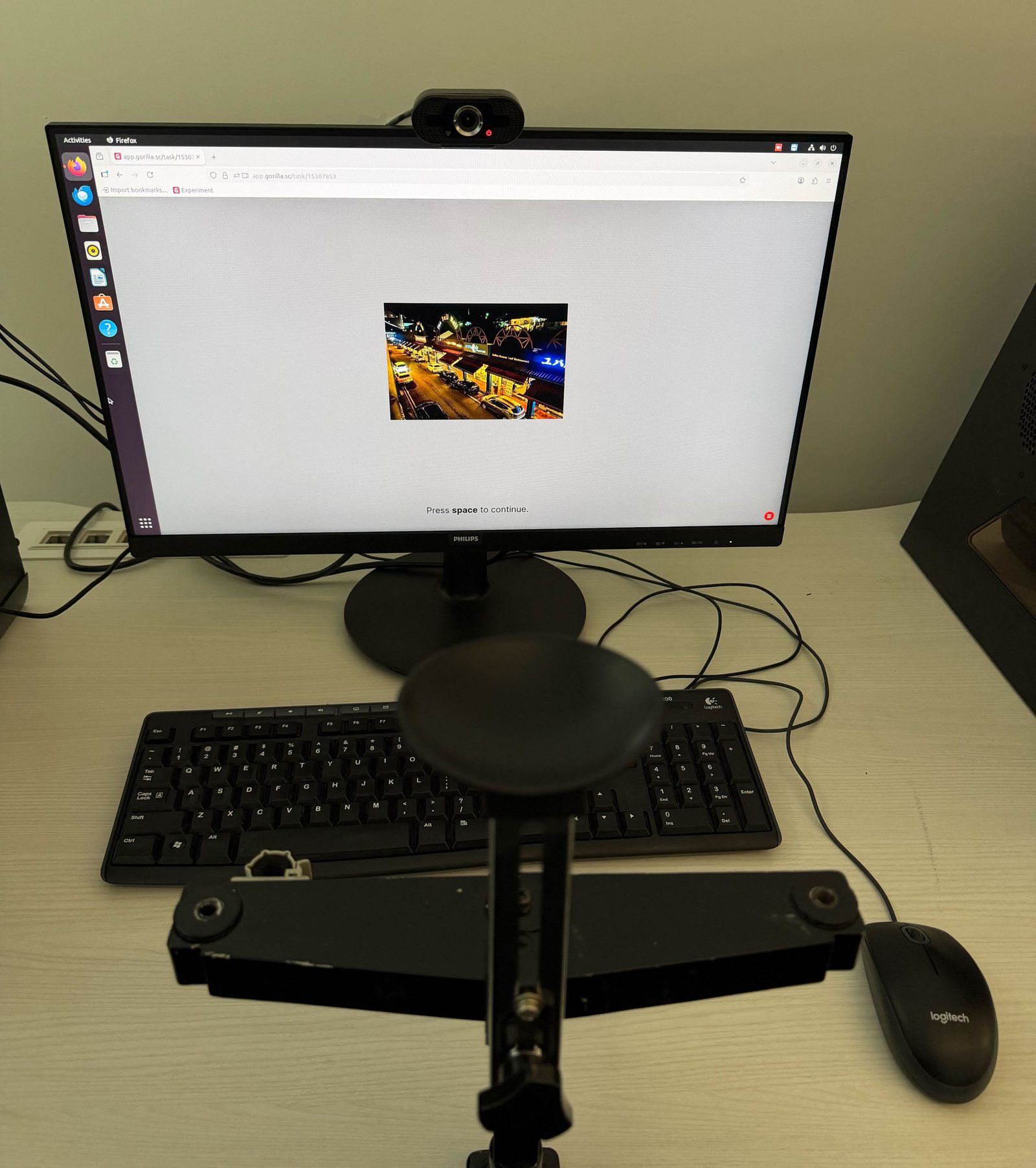}
    \caption{}
  \end{subfigure}
  \caption{\textbf{Calibration and apparatus.} (a) Online display calibration: participants resized a box to match a physical credit card, which fixes the physical stimulus size on any monitor~\cite{mayo_how_2023}. (b) Reaction-time pretest: participants pressed the space bar when a circle appeared after a random delay (50-1600~ms). (c) In-person apparatus with chin rest, calibrated monitor and webcam for eye tracking.}
  \label{fig:supp_calib}
\end{figure}

\paragraph{Setup and calibration}
Online participants were recruited through Prolific and the experiment was built in Gorilla~\cite{anwyl2020gorilla}. After consent and a demographics questionnaire, participants aligned a credit card (3.37$\times$2.13~in) with an on-screen box to resize the display (\cref{fig:supp_calib}a)~\cite{mayo_how_2023} and were instructed to sit 22~in from the screen. A reaction-time test followed (\cref{fig:supp_calib}b).

\paragraph{Image sizes}
Images were resized to fit boxes of 23.4\%$\times$31.1\% (S), 39.1\%$\times$52.1\% (M) and 62.5\%$\times$83.3\% (L) of screen width and height, preserving aspect ratio. Because the boxes are defined as fractions of the viewport, at the median viewport aspect ratio (2.0) they are about 1.5:1 rather than square; wider images are therefore limited by width and the others by height.

\begin{figure}[!ht]
  \centering
  \begin{subfigure}{0.48\linewidth}
    \centering
    \includegraphics[width=0.8\textwidth]{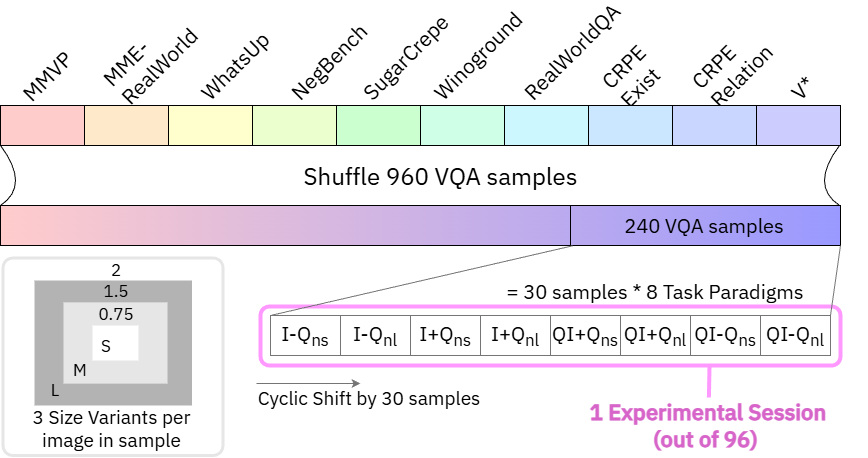}
    \caption{}
  \end{subfigure}\hfill
  \begin{subfigure}{0.48\linewidth}
    \centering
    \includegraphics[width=\textwidth]{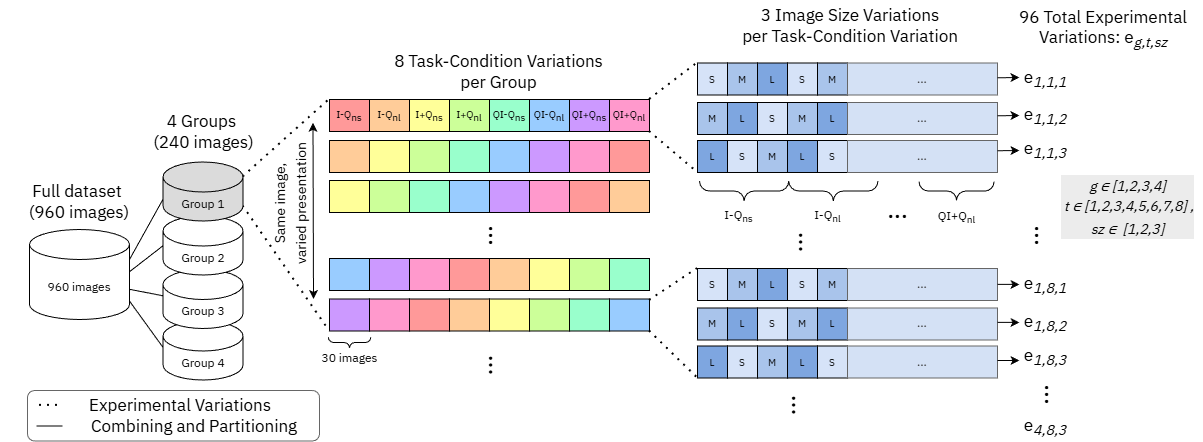}
    \caption{}
  \end{subfigure}
  \caption{\textbf{Counterbalancing.} (a) The 960 items are split into four groups of 240; each participant sees one group, with 30 items in each of the eight combinations of condition and mask duration. (b) Within a group, 30-item blocks are assigned to the eight combinations (colors) by cyclic rotation, giving eight variants per group, each crossed with a random assignment of the three image sizes (shades): $4\times8\times3=96$ variants.}
  \label{fig:supp_sorting}
\end{figure}

\paragraph{Counterbalancing}
Participants were routed at random to one of 96 session variants (\cref{fig:supp_sorting}) by three branch nodes, which choose one of four 240-item groups, one of eight block orders and one of three size assignments. When all variants are completed, every item appears in every combination of condition, mask and size, giving $960\times3\times8=23{,}040$ planned item-condition presentations. Each session had eight blocks of 30 items plus one attention check per block.

\begin{figure}[!ht]
  \centering
  \includegraphics[width=0.48\textwidth]{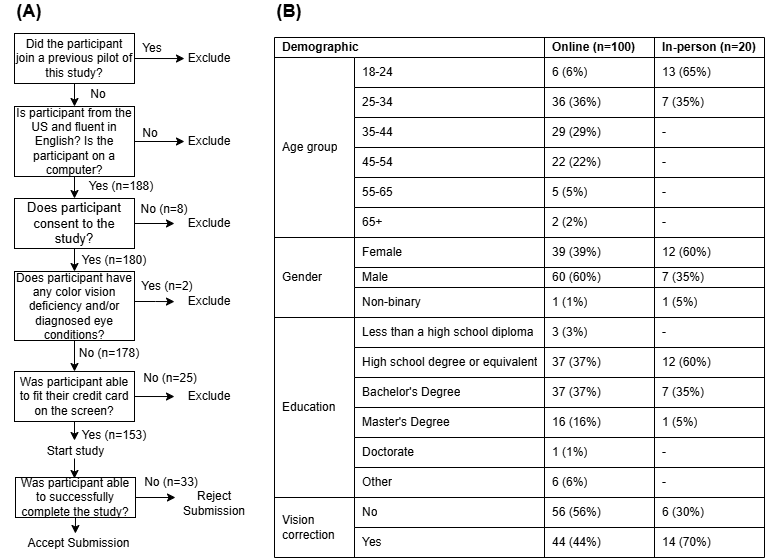}
  \caption{\textbf{Exclusions and demographics.} (A) Exclusion criteria applied to participants. (B) Demographics of the final sample.}
  \label{fig:supp_elig}
\end{figure}

\paragraph{Recruitment, exclusions and coverage}
Prolific participants were screened for US residence, English fluency and use of a computer (188 passed). Of these, 8 did not consent, 2 reported a color-vision deficiency or eye condition, and 25 could not fit the credit card on screen, so 153 started the study (\cref{fig:supp_elig}). Submissions not completed within the 140-minute limit were rejected and the slot re-advertised (33 of 153). Of the 120 accepted participants, 20 belonged to a pilot study and are not analyzed, leaving 100. Accepted participants failed at most two of the eight attention checks (84 none, 11 one, 5 two).

Because Gorilla's randomizer assigns variants at random rather than in balance, the 100 participants occupied 84 of the 96 variants (16 twice, 12 never). Since mask duration and image size are pooled in every analysis, all 3{,}840 item-condition cells are still observed, with 4-9 responses each (median 6, mean 6.2). Of the 24{,}000 online trials, 36 (0.15\%) recorded no A or B key press and are dropped, leaving 23{,}964. Approved participants were paid \$14 per hour.

\subsection{Display timing, devices and viewing distance}
\paragraph{Exposure}
Gorilla logs the realized duration of every timed screen. In the brief conditions, the image lasted a median of 83.9~ms online (5th-95th percentiles 80.3-85.1~ms). One percent of trials were one 60~Hz frame short (about 67~ms) and 0.2\% exceeded 100~ms (the longest 267~ms). In person, the median was 87.4~ms (5th-95th percentiles 69-121~ms), with 16\% of trials above 100~ms, likely because webcam gaze recording ran in the same browser. Refresh rate and dropped frames are not logged, so exposure is validated only by these screen durations, which cannot detect a missed frame within a screen.

\paragraph{Other displays}
The short mask lasted a median of 100.7~ms and the fixation 467~ms. Self-paced image views lasted a median of 1.8-2.7~s; a 50~s timeout fired on 14 trials.

\paragraph{Devices}
All 100 online participants used a desktop or laptop (Windows 82, macOS 12, ChromeOS 4, Linux 2; Chrome 76, Firefox 9, Edge 6, Safari 5, Opera 1) on 30 distinct monitor resolutions (1920$\times$1080 for 35 participants), with a median viewport width of 1{,}536 pixels. The credit-card test fixes the physical size of the display; viewing distance was instructed online and fixed by a chin rest in person. Gorilla logs neither value, so we report image sizes in inches.

\subsection{In-person study and gaze recording}
The 20 in-person sessions used a separate Gorilla experiment with WebGazer.js~\cite{papoutsaki2016webgazer} integrated for eye tracking. Sessions ran on an Ubuntu workstation with a 21.25$\times$12~in Philips monitor (1920$\times$1080, 60~Hz), a chin rest at 17~in and a Microdia Vitade AF webcam above the display (\cref{fig:supp_calib}c). Before the task, a nine-point calibration and validation (10~s per dot) was run twice and repeated until mean error fell below 4$^\circ$ of visual angle. Raw gaze coordinates were recorded at 30~Hz. Because all in-person participants used the same monitor, the credit-card step was omitted. All in-person participants saw item group~1 (240 items). They were paid in cash above local minimum wage.

\subsection{Full-option study}
\label{supp:fullopt_protocol}
Thirty-four additional Prolific participants (same screening, calibration, distance instruction and reaction test) completed 16 practice and 192 main trials in November 2026. The 192 items are 48 each from MME-RealWorld-Lite, RealWorldQA, V* and WhatsUp (\cref{tab:supp_sources}), shown with their source option sets (2-5 options, mean 4.0) at the medium size with the 200~ms mask, in one 48-item block per condition. One participant was excluded before analysis for clicking through (median reaction time 0.1~s against 1.8-8.6~s for every other participant; accuracy 30\% at a chance level of 25\%), leaving 33. Items rotated through conditions so that each has 7-9 responses per condition (mean 8.2; 6{,}336 trials). Responses are option letters scored against the source answer, and the consensus is the majority vote, as in the main study.

\subsection{Workload ratings}
After each block, participants rated the six NASA Task Load Index items~\cite{hart1988development,vriend_which_2024}: mental demand, physical demand, temporal demand, self-rated performance, effort and frustration. Results are in \cref{supp:agree}.

\FloatBarrier
%% ============================================================================ A3
\section{Additional Human Results}
\label{supp:human_results}

This section supports main \cref{sec:human}: interval estimates for the condition effects, the effects of image size and mask duration, agreement between participants, workload, and the items people miss.

\subsection{Accuracy per condition and condition contrasts}

\begin{table}[!ht]
  \centering
  \footnotesize
  \setlength{\tabcolsep}{4pt}
  \begin{tabular}{@{}lccc@{}}
    \toprule
    Condition / contrast & Accuracy or $\Delta$ (\%) & Participant 95\% & Item 95\% \\
    \midrule
    \IQm & 66.9 & [65.3, 68.7] & [65.3, 68.6] \\
    \IQp & 75.8 & [73.9, 77.5] & [74.1, 77.3] \\
    \QIQp & 81.0 & [79.4, 82.5] & [79.6, 82.4] \\
    \QIQm & 68.7 & [66.9, 70.5] & [67.1, 70.4] \\
    \midrule
    \QIQp $-$ \IQm & 14.0 & [12.5, 15.6] & [12.4, 15.7] \\
    \QIQp $-$ \QIQm & 12.2 & [10.5, 14.1] & [10.6, 13.9] \\
    \IQp $-$ \IQm & 8.8 & [7.0, 10.5] & [7.2, 10.5] \\
    \IQp $-$ \QIQm & 7.0 & [5.1, 8.8] & [5.3, 8.7] \\
    \QIQp $-$ \IQp & 5.2 & [3.7, 6.8] & [3.7, 6.6] \\
    \QIQm $-$ \IQm & 1.8 & [0.0, 3.7] & [0.2, 3.4] \\
    \bottomrule
  \end{tabular}
  \caption{\textbf{Online accuracy per condition and condition contrasts.} All 23{,}964 online trials; 95\% percentile intervals over 1{,}000 resamples of participants and, separately, of items.}
  \label{tab:supp_contrasts}
\end{table}

\paragraph{Pooled estimates}
\Cref{tab:supp_contrasts} gives the per-condition accuracy and all six contrasts over the 23{,}964 online trials, with intervals from resampling either participants or items; the two give nearly identical intervals. Every contrast involving exposure excludes zero by a wide margin. The question-order contrast under brief exposure (\QIQm minus \IQm, 1.8 points) is marginal under both.

\paragraph{How main \cref{fig:human_acc} is computed}
Each bar is the mean over participants of each participant's accuracy on the trials in that cell (setting, size or benchmark), with a 95\% $t$-interval over participants. Item dependence is not modelled in these intervals. Because in-person participants saw only item group~1, the size and benchmark panels pool 960-item online data with 240-item in-person data. The ``median of 13.3 and maximum of 24.4 points'' in main \cref{subsec:human_sensitivity} are taken over the fifteen panels (two settings, three sizes, ten benchmarks) of the \QIQp minus \IQm difference; the maximum is RealWorldQA and the minimum SugarCrepe (6.7 points).

\subsection{Image size and mask duration}
\label{supp:sizemask}

\begin{figure}[!ht]
  \centering
  \includegraphics[width=0.5\textwidth]{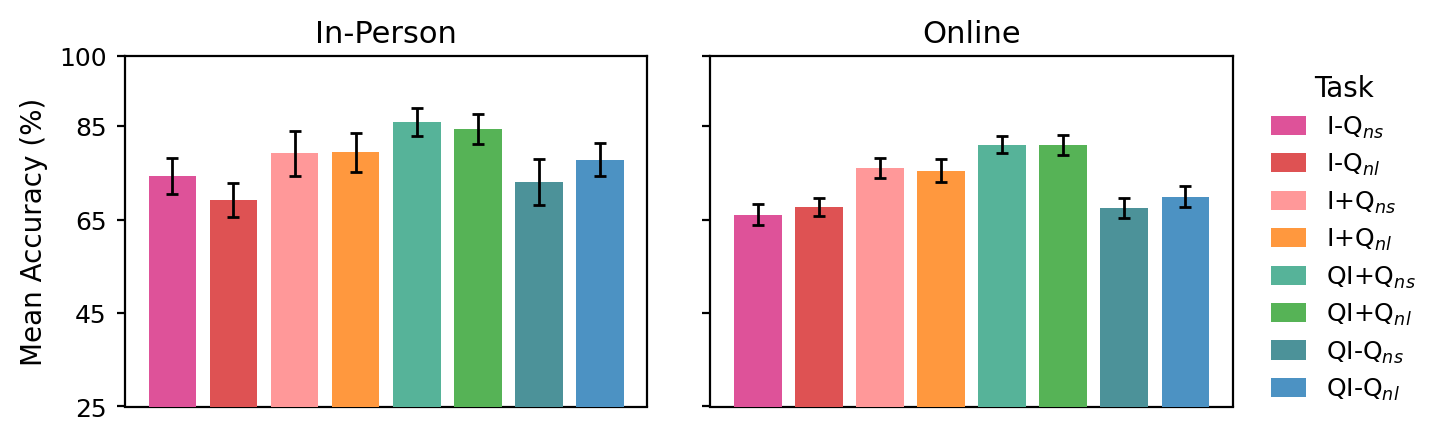}
  \caption{\textbf{No measurable effect of mask duration.} Accuracy with the short (100~ms) and long (200~ms) mask for in-person (left) and online (right) participants; 95\% CIs.}
  \label{fig:supp_mask}
\end{figure}

Both factors vary within participant.

\paragraph{Image size}
Pooled over conditions, accuracy rises from 71.7\% at the small size to 73.2\% at the medium and 74.4\% at the large size. Large minus small is $+2.7$ points (participant-bootstrap 95\% interval $[1.4, 4.2]$). Within condition it is $+3.8$ $[0.9, 6.7]$ in \IQm, $+1.2$ $[-1.2, 3.6]$ in \IQp, $+4.4$ $[2.1, 7.1]$ in \QIQp and $+1.6$ $[-1.7, 4.8]$ in \QIQm.

\paragraph{Mask duration}
Long minus short is $+0.9$ points $[-0.3, 2.1]$ pooled, and lies within $[-3.1, 4.8]$ in every condition (\cref{fig:supp_mask}).

\paragraph{Regression}
A logistic regression of trial correctness on condition, size and mask gives average marginal effects of $+13.1$ (\QIQp vs.\ \IQm), $+8.0$ (\IQp) and $+1.6$ (\QIQm) points for condition, $+2.7$ (large) and $+1.5$ (medium) for size, and $+0.9$ for mask. Their 95\% intervals under participant-clustered standard errors are $[11.8, 14.5]$, $[6.4, 9.6]$, $[0.0, 3.2]$, $[1.4, 4.1]$, $[0.2, 2.8]$ and $[-0.4, 2.1]$; item-clustered intervals are within 0.3 points. A mixed logit with random intercepts for participant and item gives the same effects to within 0.1 points. Neither the condition-by-size nor the condition-by-mask interaction is significant (Wald $p=0.19$; likelihood-ratio $p=0.21$). Image size therefore has a small additive effect, an order of magnitude below the exposure effect, and mask duration none we can detect; both are pooled in every analysis.

\subsection{Agreement between participants and workload}
\label{supp:agree}

\begin{figure*}[!ht]
  \centering
  % panels d,e of the human-accuracy figure
  \overlayfig{\includegraphics[width=0.9\textwidth,viewport=30 18 1560 708,clip]{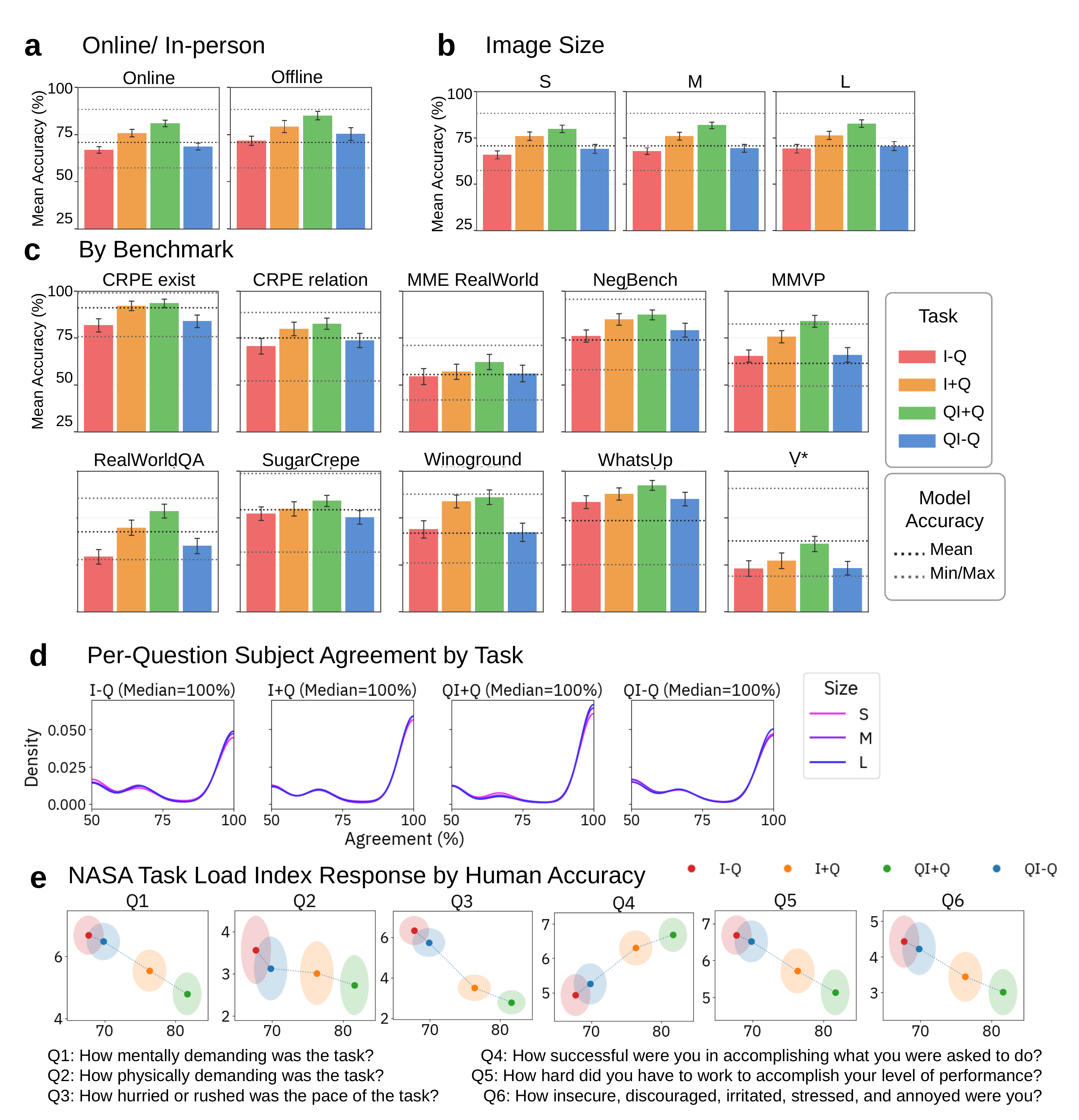}}{%
    \relabel{0.0}{0.91}{0.035}{1.0}{A}
    \relabel{0.0}{0.45}{0.035}{0.56}{B}}
  \caption{\textbf{Participant agreement and workload.} (A) Per-item agreement between participants who saw the same item in the same condition, \emph{split by image size}. This leaves one to four responses per cell, so the medians of 100\% reflect the small count rather than high agreement; see the text for unsplit values. (B) NASA Task Load Index ratings after each block, one panel per item (Q1-Q6). Each point is one condition's mean rating against its mean accuracy; ellipses show spread across participants.}
  \label{fig:supp_agree}
\end{figure*}

\paragraph{Agreement}
For each item and condition we take the share of online responses that match the modal response (ties count 50\%). Over the 3{,}840 cells (4-9 responses, median 6), the median is 75.0\% in \IQm, 83.3\% in \IQp, 85.7\% in \QIQp and 80.0\% in \QIQm (means 76.5, 81.9, 85.6 and 78.0\%). Responses are unanimous in 21.6, 34.6, 44.8 and 24.2\% of cells and tied in 7.9, 7.0, 3.5 and 5.8\%. \Cref{fig:supp_agree}A shows the same quantity split further by image size.

\paragraph{Workload}
\Cref{fig:supp_agree}B shows that reported load falls as accuracy rises across conditions for the five load items, while self-rated success (Q4) rises. The hardest condition (\IQm) is also rated the most demanding.

\subsection{Items that people miss}
\label{supp:human_misses}

This section expands main \cref{subsec:human_misses} and \cref{fig:asym}.

\paragraph{Model misses and human misses}
Against the \QIQp consensus, the number of items a model misses but people get right varies widely: 28 for Gemini-3.8-Flash and 290-353 for the question-agnostic models. The reverse, items the model gets right but people miss, numbers 32-71 for every model (main \cref{fig:asym}A). Pooled over the eleven strong autoregressive models, 76\% of these human misses come from V*, MME-RealWorld or RealWorldQA, against 30\% of all items (main \cref{fig:asym}B).

\paragraph{Properties of the human misses}
\QIQp participants agree with each other on only 70\% of these items, against 88\% on items both get right, and knowing the question does not help on them (mean gain from \IQm to \QIQp of $-0.18$, against $+0.14$ over all items). Eighteen items are answered correctly by all eleven strong models while at most half of \QIQp participants get them; nine are from V*. Image size within our 2.5-6.7~in range changed accuracy by only 2.7 points (\cref{supp:sizemask}), consistent with targets that remain too small at every size we tested; a zoom condition would test this directly.

\paragraph{Items that defeat models}
Model failures on items people find easy are heterogeneous. Of the 518 items that at least 85\% of \QIQp participants answer correctly, 261 are solved by all eleven strong models and only two are missed by nine or more.

\FloatBarrier
%% ============================================================================ A4
\section{Model Evaluation}
\label{supp:models}

This section expands main \cref{subsec:models}: prompts, checkpoints, decoding, how results are aggregated over templates, excluded candidates, and per-benchmark accuracy.

\subsection{Prompt templates}

\begin{table}[!ht]
  \centering
  \caption{Prompt templates for autoregressive VLMs.}
  \label{tab:supp_ar}
  \small
  \begin{tcolorbox}
  base = ``Question: \{question\}\textbackslash nOptions: \{options\}\textbackslash n\textbackslash n''

  1: base + ``Please select the correct option. Respond ONLY with the single letter corresponding to the correct answer (e.g., A, B). Do not add punctuation, spaces, or explanation.''

  2: base + ``Respond ONLY with the single letter corresponding to the correct answer (e.g., A, B). Do not add punctuation, spaces, or explanation.''

  3: base + ``Respond ONLY with the single letter corresponding to the correct answer (e.g., A, B).''

  4: base + ``Your response should be only one letter (e.g., A, B).''

  5: base + ``(out of options, please select ONLY ONE correct option. No explanation needed.)''
  \end{tcolorbox}
\end{table}

\begin{table}[!ht]
  \centering
  \caption{Prompt templates for question-agnostic models (contrastive dual encoders and BLIP-ITM). Each option is scored by image-text similarity or matching score; \texttt{letters = sorted(options.keys())}.}
  \label{tab:supp_dual}
  \small
  \begin{tcolorbox}
  1: [f``\{question\} \{options[L]\}'' for L in letters]

  2: [f``Question:\{question\} \textbackslash n Answer: \{options[L]\}'' for L in letters]

  3: [f``USER: \{question\} \textbackslash n ASSISTANT:\{options[L]\}'' for L in letters]

  4: [f``\{options[L]\} \{question\}'' for L in letters]

  5: [f``The answer to \{question\} is \{options[L]\}'' for L in letters]
  \end{tcolorbox}
\end{table}

Autoregressive models were prompted with the instructions in \cref{tab:supp_ar}. Question-agnostic models scored the image against each option rendered with the caption-style templates in \cref{tab:supp_dual} and chose the higher-scoring option. Some models did not follow the single-letter instruction: the two Gemini models and \llamathree produced full sentences for templates 4 and 5, and the four GPT models for template 5. We therefore use the first three templates for the former and the first four for the latter.

\subsection{Checkpoints, decoding and preprocessing}

\begin{table}[!ht]
  \centering
  \footnotesize
  \setlength{\tabcolsep}{3pt}
  \begin{tabular}{@{}lll@{}}
    \toprule
    Model & Identifier & Source \\
    \midrule
    LLaMA-3.2-11B-Vision & \texttt{meta-llama/Llama-3.2-11B-Vision-Instruct} & HF \\
    LLaVA-1.5-7B / 13B & \texttt{llava-hf/llava-1.5-7b-hf}, \texttt{-13b-hf} & HF \\
    PaliGemma-3B & \texttt{google/paligemma-3b-mix-224} & HF \\
    Qwen2.5-VL-3B / 7B & \texttt{Qwen/Qwen2.5-VL-3B-Instruct}, \texttt{-7B-Instruct} & HF \\
    InternVL2.5-1B & \texttt{OpenGVLab/InternVL2\_5-1B} & HF \\
    nanoLLaVA & \texttt{qnguyen3/nanoLLaVA} & HF \\
    Gemini-2.5-Flash-Lite & \texttt{gemini-2.5-flash-lite} & Google API \\
    Gemini-3.8-Flash & \texttt{gemini-3.8-flash} & Google API \\
    GPT-4o-mini / 4.1-mini / 4.1 & \texttt{gpt-4o-mini}, \texttt{gpt-4.1-mini}, \texttt{gpt-4.1} & OpenAI API \\
    GPT-6-Luna & \texttt{GPT-6-Luna} & OpenAI API \\
    CLIP ViT-B/16, B/32 & \texttt{ViT-B-16}, \texttt{ViT-B-32} (\texttt{openai}) & open\_clip \\
    CLIP ViT-L/14-336 & \texttt{ViT-L-14-336} (\texttt{openai}) & open\_clip \\
    SigLIP & \texttt{ViT-B-16-SigLIP} (\texttt{webli}) & open\_clip \\
    CoCa & \texttt{coca\_ViT-L-14} (\texttt{mscoco\_finetuned\_laion2B-s13B-b90k}) & open\_clip \\
    BLIP-ITC / BLIP-ITM & \texttt{Salesforce/blip-itm-large-coco} & HF \\
    \bottomrule
  \end{tabular}
  \caption{\textbf{Model identifiers.} Hugging Face (HF) repositories, open\_clip architecture (pretrained tag), or API model names as evaluated.}
  \label{tab:supp_models}
\end{table}

\paragraph{Checkpoints and decoding}
\Cref{tab:supp_models} lists every checkpoint and API identifier. Local autoregressive models decode greedily (\texttt{do\_sample=False}, at most 8 new tokens) in bfloat16, or float16 on GPUs below compute capability 8. API models are called with temperature 0 and a system message: GPT with at most 10 tokens and images sent as base64 JPEG at the default detail setting, Gemini with at most 50 output tokens. A twenty-first model, the contrastive \clipl encoder used inside LLaVA-1.5, appears only in the shared-encoder comparison (\cref{tab:backbone}); every other analysis uses the 20 main models unless a subset is stated.

\paragraph{Image preprocessing}
Each model receives images through its own processor at the default resolution: Qwen2.5-VL with default pixel limits, LLaMA-3.2 with up to four $560^2$ tiles, InternVL2.5 with up to twelve $448^2$ tiles plus a thumbnail, LLaVA-1.5 at $336^2$, PaliGemma at $224^2$ and nanoLLaVA at $384^2$. open\_clip models use their default resize and center crop ($224^2$, or $336^2$ for ViT-L/14-336) and BLIP its processor's $384^2$ input.

\paragraph{Answer parsing}
The answer letter of a local model is parsed from the generated text after stripping punctuation (the first character if it is A or B, otherwise a stand-alone A or B). API responses are compared to the correct letter after stripping punctuation and whitespace. Question-agnostic models score each option by the cosine similarity between the L2-normalized image embedding and the text embedding of the templated option and take the higher one; BLIP-ITM uses its matching-head probability instead.

\paragraph{Compute and software}
The main evaluation ran on 2-3 March 2026 (NVIDIA H100 and H200); the additional models, resolution sweeps, question-guided crops and original-option reruns on 17-25 October 2026 (L40S and P100); and D-RISE on 18-24 October 2026. Model code ran under Python 3.9 with \texttt{transformers}, \texttt{open\_clip} and the \texttt{openai} and \texttt{google.generativeai} SDKs; the analysis under Python 3.11 with numpy 1.26, pandas 1.5 and statsmodels 0.14.

\subsection{Aggregation over templates, ties and invalid outputs}
\label{supp:aggregation}

\paragraph{Aggregation}
For each valid template $t$, a model has a binary correctness vector $y_{m,i}^{(t)}$ over the 960 items. Observed disagreement with a human condition is computed per template and averaged,
$\bar{d}=\frac1T\sum_t\frac1N\sum_i[y_{m,i}^{(t)}\neq\hat{y}_{h,i}]$,
which equals the disagreement between the consensus and the template-averaged correctness $\bar{y}_{m,i}=\frac1T\sum_t y_{m,i}^{(t)}$. Chance disagreement is simulated once, with $R=100$ random responders at the template-averaged accuracy $\bar{a}_m=\frac1N\sum_i\bar{y}_{m,i}$. The ratio in main \cref{fig:families} is $\bar{d}/(1-\bar{p}_e)$. Its error bars are the SD over templates of the per-template observed disagreement divided by the same chance term, so they carry no chance-simulation spread. The accuracy shown in parentheses in main \cref{fig:families} and in \cref{tab:supp_boot} is $\bar{a}_m$.

\paragraph{Which estimator each analysis uses}
The item bootstrap (\cref{tab:supp_boot}), the shared-encoder comparison (\cref{tab:backbone}), the difficulty-stratified null (\cref{fig:supp_e19}), the participant bootstrap (\cref{fig:supp_e20}) and the mixed model use $\bar{y}_{m,i}$ with the analytic chance term $\sum_i[\bar{a}_m(1-\hat{y}_{h,i})+(1-\bar{a}_m)\hat{y}_{h,i}]$. Cohen's $\kappa$ (\cref{fig:supp_kappa_main}) is computed per template and reported as mean $\pm$ SD over templates. Only the reliability correction (\cref{fig:supp_e3}) uses a majority vote over templates (ties scored correct). The resolution sweeps use three templates and the original-option reruns five.

\paragraph{Template ties}
Because correctness is averaged over templates, a 2-2 split among the four GPT templates counts one half. This happens on 13, 35 and 27 items for GPT-4.1, GPT-4.1-mini and GPT-4o-mini. Scoring these items as zero lowers accuracy by 0.7, 1.8 and 1.4 points and changes no best match. (Under a majority vote with ties scored correct, accuracy would be 87.6\%, 85.9\% and 73.6\%.)

\paragraph{Invalid outputs}
Outputs that are not a single option letter (full sentences, refusals, ``AB'' or a letter outside the option set) are scored incorrect for that template: 126 of the 79{,}680 template responses of the original 18 models, 48 of them from Gemini-2.5-Flash-Lite. One MMVP image failed to load for the six question-agnostic models, InternVL2.5-1B and nanoLLaVA and is likewise scored incorrect. Neither choice changes a best match.

\subsection{Excluded candidate models}
Two further candidates were run with the same templates and greedy decoding and then excluded. InstructBLIP (Vicuna-7B) answered at chance (52\%) under every template, which may reflect instruction-following or scoring problems that its result alone cannot identify. BLIP-2 (Flan-T5-XL) decodes with a language model but reads the image through a Q-Former whose visual tokens are computed before the question is seen, so it fits neither family cleanly. For the record, it scored 68\% and matched \IQm and \IQp equally.

\subsection{Accuracy per benchmark}

\begin{figure*}[!ht]
  \centering
  \includegraphics[width=\textwidth]{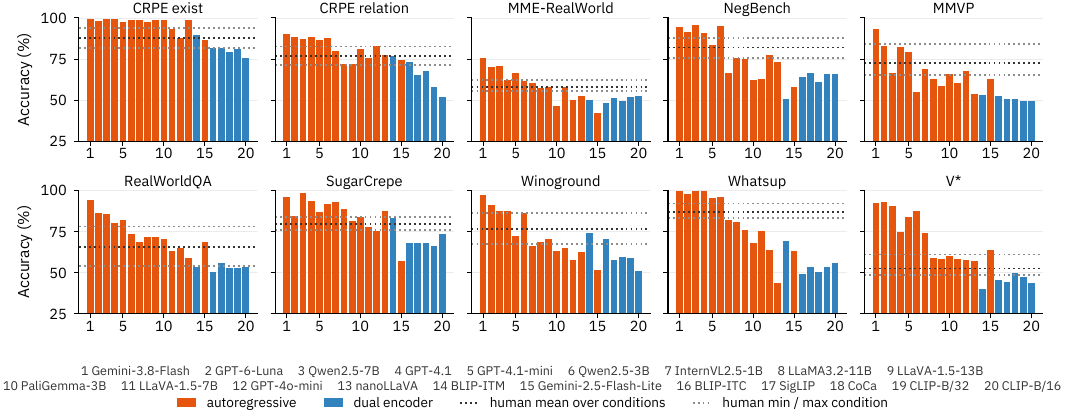}
  \caption{\textbf{Model accuracy (\%) on each benchmark.} Models are numbered by overall accuracy (key below); autoregressive models in orange, dual encoders in blue. Dotted lines mark the human mean and the minimum and maximum over the four conditions.}
  \label{fig:supp_modelacc}
\end{figure*}

\Cref{fig:supp_modelacc} reports per-benchmark accuracy for all 20 models alongside the human range across conditions.

\FloatBarrier
%% ============================================================================ A5
\section{Agreement Metrics}
\label{supp:agreement}

This section gives the full definitions behind main \cref{subsec:metric,subsec:families}: the human consensus, Cohen's $\kappa$ and the disagreement ratio, the complete per-model results, and checks of how certain the best matches are.

\subsection{Human consensus and the tie rule}
\label{supp:consensus}
Only online participants contribute to the item-level consensus; in-person participants were too few to cover all items evenly. The consensus for an item and condition is \emph{correct} when at least half of its 4-9 responses are correct, so tied votes count as correct. Ties occur in 233 of the 3{,}840 cells (6.1\%: 76 in \IQm, 67 in \IQp, 34 in \QIQp and 56 in \QIQm).

Scoring ties as incorrect leaves every best match unchanged except \gptfouro (to \IQp) and GPT-6-Luna (to \QIQp); dropping tied cells instead moves \llamathree, \gemini and \gptfouro from \QIQp to \IQp. The count of eight autoregressive models at \QIQp in main \cref{subsec:families} is therefore specific to the tie rule (eight or five under the two alternatives), while every dual encoder keeps its glimpse match under all three. The vote-free partial correlation of \cref{supp:reliability} does not depend on it.

Because agreement is computed on correctness, it measures shared success and failure. Chance correction accounts for each observer's overall accuracy but not for shared item difficulty, which the controls in \cref{supp:controls} address.

\subsection{Cohen's $\kappa$ and the disagreement ratio}

\paragraph{Cohen's $\kappa$}
With $y_{m,i}$ the model's correctness and $\hat{y}_{h,i}$ the consensus, observed agreement is $p_o=\frac1N\sum_i[y_{m,i}=\hat{y}_{h,i}]$ and chance agreement $p_e=\sum_{k}P(M{=}k)P(H{=}k)$ from the two marginals. Then $\kappa=(p_o-p_e)/(1-p_e)$~\cite{cohen1960coefficient,mchugh2012interrater}: 1 for identical error patterns, 0 for chance and negative below chance. We follow Landis and Koch~\cite{landis1977measurement} for interpretation.

\paragraph{Disagreement ratio}
For the family analysis we estimate chance empirically. For each model we generate $R=100$ simulated responders with the model's exact accuracy but uniformly random error placement,
\begin{equation}
\bar{p}_e=\frac{1}{RN}\sum_{r=1}^{R}\sum_{i=1}^{N}[\,y_{r,i}=\hat{y}_{h,i}\,],
\end{equation}
and divide observed disagreement $1-p_o$ by chance disagreement $1-\bar{p}_e$. For large $R$ this ratio approximates $1-\kappa$.

\begin{figure*}[!ht]
  \centering
  \includegraphics[width=0.95\textwidth]{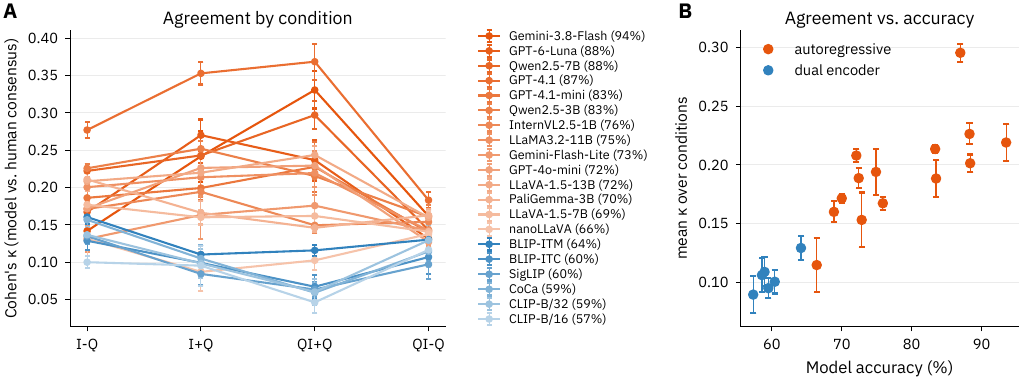}
  \caption{\textbf{Cohen's $\kappa$ between models and humans.} (A) $\kappa$ between each model and the consensus in each condition; mean and SD across templates. Autoregressive models in coral, dual encoders in periwinkle; darker lines for more accurate models. (B) Agreement above chance, averaged over conditions, against model accuracy.}
  \label{fig:supp_kappa_main}
\end{figure*}

\begin{figure*}[!ht]
  \centering
  \includegraphics[width=0.95\textwidth]{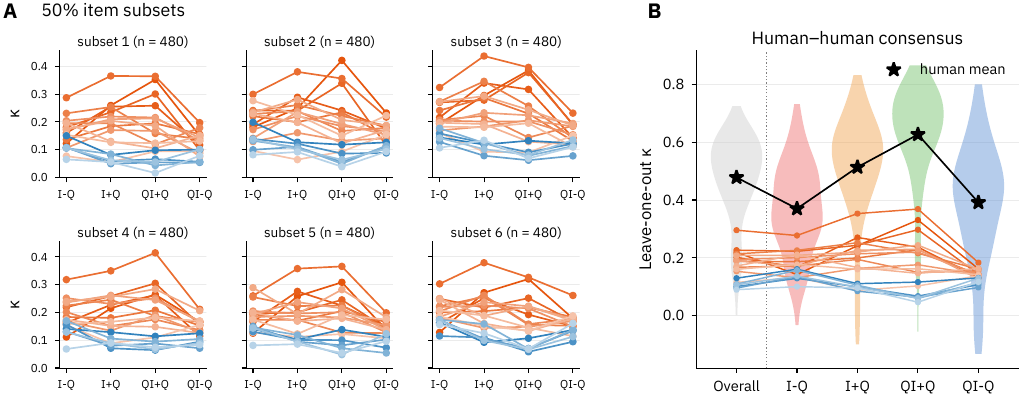}
  \caption{\textbf{Robustness of $\kappa$ and a human-human reference.} (A) $\kappa$ recomputed on six random halves of the items (colors as in \cref{fig:supp_kappa_main}). (B) Leave-one-out human $\kappa$: each online participant against the majority of the others on the same items, overall and per condition (violins; stars mark means), with model-human $\kappa$ overlaid. Mean human-human $\kappa$ is 0.37, 0.51, 0.63 and 0.39 in \IQm, \IQp, \QIQp and \QIQm.}
  \label{fig:supp_kappa}
\end{figure*}

\paragraph{Results}
\Cref{fig:supp_kappa_main} shows $\kappa$ for all 20 models. The stronger autoregressive models tend to peak in \QIQp and the question-agnostic models in a brief condition, and this holds on random halves of the items (\cref{fig:supp_kappa}A). The best model-condition pairs reach about $\kappa=0.4$, whereas a participant agrees with the remaining participants at 0.63 in \QIQp and 0.51 in \IQp (0.37 and 0.39 in the brief conditions; \cref{fig:supp_kappa}B). Disagreement among people therefore does not explain the modest model agreement in the self-paced conditions.

\subsection{Full disagreement profiles}

\begin{figure*}[!ht]
  \centering
  \includegraphics[width=0.95\textwidth]{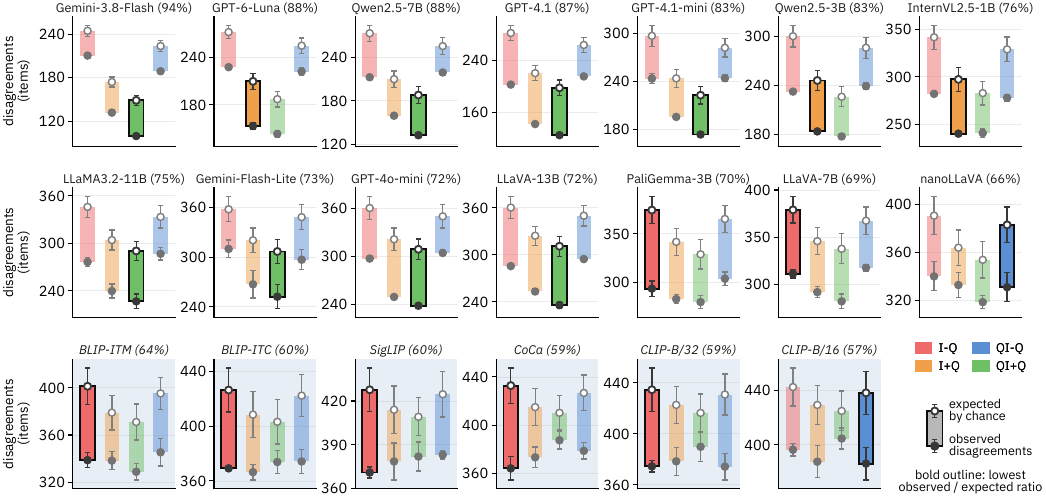}
  \caption{\textbf{Observed and chance disagreements per model.} For each model and human condition, the number of items on which model and consensus disagree: expected by chance at the model's accuracy (open circle) and observed (filled circle). Main \cref{fig:families} plots their ratio. Bold outline: best match.}
  \label{fig:supp_counts}
\end{figure*}

\Cref{fig:supp_counts} shows the observed and chance disagreements separately. For any model above 50\% accuracy, chance disagreement is largest with the least accurate reference, so raw counts would favor the more accurate conditions. The ratio corrects for this. \siglip\ is an example: under random error placement it would disagree most with the \IQm consensus (about 430 items) and least with the \QIQp consensus (about 410), yet it observably disagrees least with \IQm (about 370), giving its lowest ratio, 0.87, and its best match. This reversal indicates that \siglip\ and glimpse viewers fail on overlapping items more than accuracy alone predicts; the difficulty-stratified null (\cref{supp:null}) tests how much of this overlap shared item difficulty explains.

\subsection{Permutation test and held-out benchmarks}
\label{supp:permutation}

\begin{figure*}[!ht]
  \centering
  \includegraphics[width=0.95\textwidth]{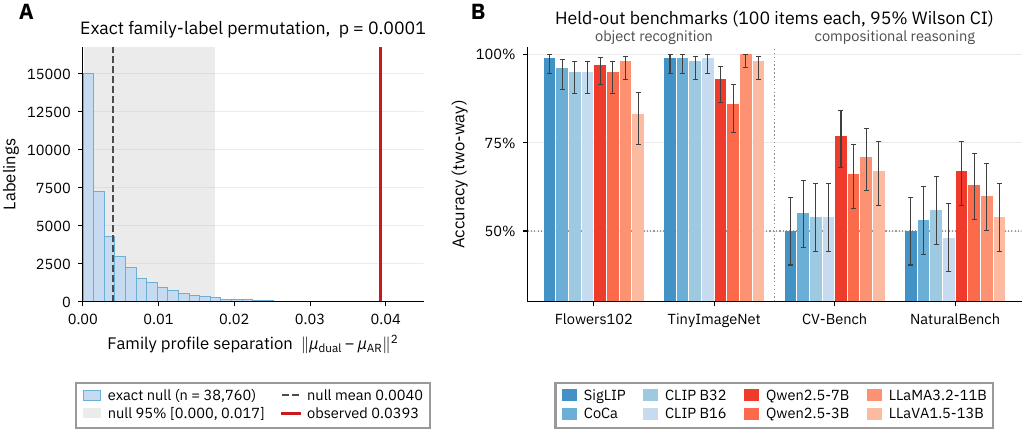}
  \caption{\textbf{Permutation test and held-out benchmarks.} (A) Exact null distribution of the separation between family mean profiles over all 38{,}760 labelings of the 20 models; the observed separation (0.039) lies far outside the null's 95\% range ($p=0.0001$; lineage-aware versions in \cref{supp:permutation}). (B) Two-way accuracy of four dual encoders (blue) and four autoregressive models (red) on 100 items from each of four held-out benchmarks, with 95\% Wilson intervals; dotted line: chance. Dual encoders match or exceed autoregressive models on object recognition (Flowers102, TinyImageNet) and fall near chance on compositional reasoning (CV-Bench, NaturalBench).}
  \label{fig:supp_dissociation_bc}
\end{figure*}

\paragraph{Permutation test}
For each model we form the four-vector of disagreement ratios over \IQm, \IQp, \QIQp and \QIQm. The family mean profiles $\mu_{\mathrm{dual}}$ and $\mu_{\mathrm{AR}}$ are compared by $\|\mu_{\mathrm{dual}}-\mu_{\mathrm{AR}}\|^2$ against the exact null over all $\binom{20}{6}=38{,}760$ labelings that preserve group sizes (6 question-agnostic, 14 autoregressive). The $p$-value is the fraction of labelings with a distance at least as large as observed. This statistic does not assume a preferred condition for either family. The observed distance is 0.0393 against a null mean of 0.0040 and 95\% range $[0.000, 0.017]$; 3 of 38{,}760 labelings reach it ($p=0.0001$; \cref{fig:supp_dissociation_bc}A). Because related models (\eg the four GPT models) are not independent, we also ran two lineage-aware versions with the lineage groups of \cref{supp:mixed}. Permuting family labels over whole lineages (12 lineages, 495 labelings) gives a null mean of 0.007 and 95\% range [0.000, 0.027]; the observed split is the most extreme of all labelings ($p=0.002$, the smallest attainable). Averaging each lineage into a single profile gives an observed distance of 0.030 against a null mean of 0.005 and range [0.000, 0.021] ($p=0.006$).

\paragraph{Held-out benchmarks}
To test whether the families' accuracy difference extends beyond our item pool, eight models (\siglip, \coca, \clipthirtytwo, \clipsixteen; \qwenseven, \qwenthree, \llamathree, \llavathirteen) answered 100 two-way items from each of four further benchmarks, with a single prompt template and scored as in the main study. For Flowers102 and TinyImageNet, the question asks which type of flower or main object is shown, with the true class against one random other class. For CV-Bench, items are drawn at random from all four tasks (count, relation, depth and distance), and each multi-way item is reduced to the correct option and one random distractor. For NaturalBench, 25 yes/no groups are expanded to all four image-question pairs and scored per item. We report accuracy with 95\% Wilson intervals (about $\pm5$ points near 95\% accuracy and $\pm10$ near 50\%).

Pooled by family, dual encoders reach 96\% and 99\% on the two recognition benchmarks but only 53\% and 52\% on the two compositional ones, against 93\% and 94\% versus 70\% and 61\% for the autoregressive models (\cref{fig:supp_dissociation_bc}B). Dual encoders thus match or exceed autoregressive models on object recognition and fall to near chance on compositional reasoning, consistent with the known relational limits of contrastive representations~\cite{kang_is_2025,yuksekgonul_when_2023}.

\subsection{Item bootstrap of the best match}
\label{supp:bootstrap}

\begin{table*}[!ht]
  \centering
  \caption{\textbf{Bootstrap of the best-matched condition.} Probability over 1{,}000 item resamples that each condition has the lowest disagreement ratio; the match on all items; and the resemblance index with its 95\% interval. Question-agnostic models in italics. The lower block gives the resolution-sweep settings of \cref{supp:resolution}.}
  \label{tab:supp_boot}
  \small
  \begin{tabular}{@{}lrccccccc@{}}
\toprule
Model & Acc.\ (\%) & P(I$-$Q) & P(I$+$Q) & P(QI$-$Q) & P(QI$+$Q) & Match & Index & 95\% interval \\
\midrule
Gemini-3.8-Flash & 93.5 & 0.00 & 0.03 & 0.00 & 0.97 & QI$+$Q & $+0.188$ & $[+0.104,\,+0.270]$ \\
GPT-6-Luna & 88.4 & 0.00 & 0.77 & 0.00 & 0.23 & I$+$Q & $+0.069$ & $[+0.000,\,+0.141]$ \\
Qwen2.5-7B & 88.3 & 0.04 & 0.10 & 0.00 & 0.86 & QI$+$Q & $+0.076$ & $[-0.027,\,+0.173]$ \\
GPT-4.1 & 87.0 & 0.01 & 0.39 & 0.00 & 0.61 & QI$+$Q & $+0.092$ & $[+0.001,\,+0.190]$ \\
GPT-4.1-mini & 83.5 & 0.11 & 0.21 & 0.00 & 0.68 & QI$+$Q & $+0.041$ & $[-0.042,\,+0.123]$ \\
Qwen2.5-3B & 83.4 & 0.24 & 0.63 & 0.00 & 0.13 & I$+$Q & $-0.012$ & $[-0.096,\,+0.070]$ \\
InternVL2.5-1B & 75.9 & 0.20 & 0.67 & 0.08 & 0.06 & I$+$Q & $-0.020$ & $[-0.098,\,+0.062]$ \\
LLaMA3.2-11B & 74.9 & 0.20 & 0.35 & 0.00 & 0.45 & QI$+$Q & $+0.019$ & $[-0.062,\,+0.098]$ \\
Gemini-2.5-Flash-Lite & 72.8 & 0.04 & 0.29 & 0.10 & 0.56 & QI$+$Q & $+0.044$ & $[-0.027,\,+0.118]$ \\
GPT-4o-mini & 72.4 & 0.03 & 0.42 & 0.00 & 0.55 & QI$+$Q & $+0.058$ & $[-0.018,\,+0.139]$ \\
LLaVA-1.5-13B & 72.1 & 0.14 & 0.21 & 0.00 & 0.64 & QI$+$Q & $+0.034$ & $[-0.046,\,+0.113]$ \\
PaliGemma-3B & 70.0 & 0.82 & 0.08 & 0.07 & 0.03 & I$-$Q & $-0.064$ & $[-0.133,\,+0.009]$ \\
LLaVA-1.5-7B & 68.9 & 0.51 & 0.21 & 0.05 & 0.23 & I$-$Q & $-0.015$ & $[-0.084,\,+0.057]$ \\
nanoLLaVA & 66.4 & 0.38 & 0.02 & 0.52 & 0.08 & QI$-$Q & $-0.029$ & $[-0.095,\,+0.031]$ \\
\textit{BLIP-ITM} & 64.2 & 0.79 & 0.04 & 0.13 & 0.04 & I$-$Q & $-0.045$ & $[-0.101,\,+0.022]$ \\
\textit{BLIP-ITC} & 60.4 & 0.71 & 0.11 & 0.17 & 0.01 & I$-$Q & $-0.063$ & $[-0.121,\,-0.006]$ \\
\textit{SigLIP} & 59.5 & 0.91 & 0.03 & 0.06 & 0.00 & I$-$Q & $-0.071$ & $[-0.121,\,-0.020]$ \\
\textit{CoCa} & 59.0 & 0.92 & 0.03 & 0.04 & 0.00 & I$-$Q & $-0.096$ & $[-0.147,\,-0.046]$ \\
\textit{CLIP-B/32} & 58.6 & 0.59 & 0.04 & 0.37 & 0.00 & I$-$Q & $-0.076$ & $[-0.127,\,-0.028]$ \\
\textit{CLIP-B/16} & 57.3 & 0.22 & 0.14 & 0.65 & 0.00 & QI$-$Q & $-0.054$ & $[-0.104,\,-0.006]$ \\
\midrule
Qwen2.5-7B @ 1280 tok & 87.5 & 0.01 & 0.02 & 0.00 & 0.97 & QI$+$Q & $+0.123$ & $[+0.021,\,+0.219]$ \\
Qwen2.5-7B @ 256 tok & 83.4 & 0.14 & 0.37 & 0.01 & 0.48 & QI$+$Q & $+0.030$ & $[-0.065,\,+0.124]$ \\
Qwen2.5-7B @ 64 tok & 80.9 & 0.11 & 0.83 & 0.01 & 0.04 & I$+$Q & $-0.016$ & $[-0.101,\,+0.075]$ \\
Qwen2.5-7B @ 16 tok & 76.1 & 0.07 & 0.60 & 0.25 & 0.07 & I$+$Q & $-0.001$ & $[-0.083,\,+0.075]$ \\
InternVL2.5-1B @ 12 tiles & 75.7 & 0.25 & 0.62 & 0.10 & 0.03 & I$+$Q & $-0.033$ & $[-0.114,\,+0.048]$ \\
LLaMA3.2-11B @ 4 tiles & 75.1 & 0.16 & 0.30 & 0.01 & 0.54 & QI$+$Q & $+0.028$ & $[-0.054,\,+0.109]$ \\
LLaMA3.2-11B @ 2 tiles & 75.0 & 0.21 & 0.08 & 0.00 & 0.70 & QI$+$Q & $+0.028$ & $[-0.043,\,+0.107]$ \\
InternVL2.5-1B @ 4 tiles & 73.6 & 0.46 & 0.49 & 0.00 & 0.04 & I$+$Q & $-0.044$ & $[-0.120,\,+0.035]$ \\
LLaMA3.2-11B @ 1 tiles & 73.1 & 0.36 & 0.49 & 0.01 & 0.14 & I$+$Q & $-0.013$ & $[-0.090,\,+0.065]$ \\
InternVL2.5-1B @ 1 tiles & 72.9 & 0.54 & 0.42 & 0.01 & 0.03 & I$-$Q & $-0.052$ & $[-0.124,\,+0.030]$ \\
\bottomrule
\end{tabular}
\end{table*}

For every model we resampled the 960 items with replacement 1{,}000 times and recomputed the four ratios, with chance computed analytically at the model's accuracy on each resample. \Cref{tab:supp_boot} lists the probability that each condition is the best match and the 95\% interval of the index.

Only 5 of 20 best matches have bootstrap support of 0.8 or more (Gemini-3.8-Flash, \qwenseven, \gemma, \siglip, \coca). This is why the main text relies on the continuous index and on family-level summaries. The index is rank-correlated with accuracy at 0.88 over all models and 0.77 within the autoregressive models. No question-agnostic model has a \QIQp probability above 0.04.

\subsection{Reference reliability and alternative statistics}
\label{supp:reliability}

\begin{figure*}[!ht]
  \centering
  \includegraphics[width=0.95\textwidth]{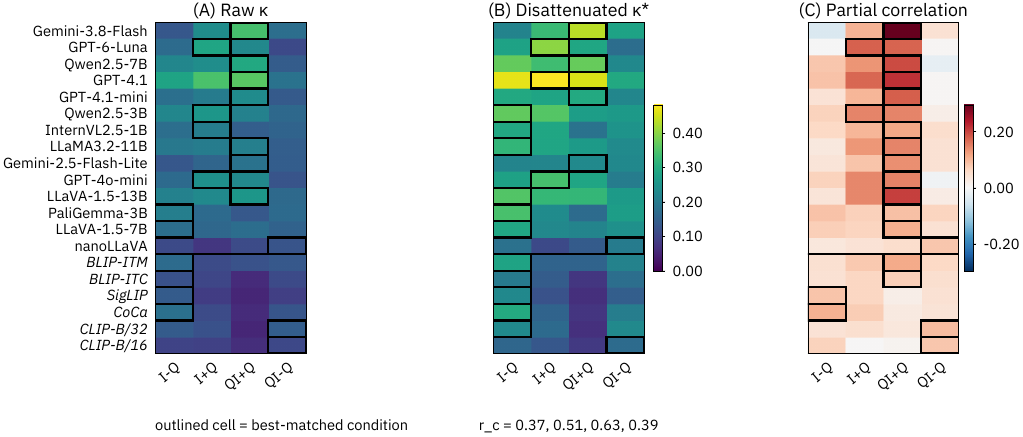}
  \caption{\textbf{Three agreement statistics.} (A) Raw Cohen's $\kappa$. (B) Reliability-rescaled $\kappa^{*}=\kappa/\sqrt{r_c}$, with $r_c$ the mean leave-one-out human $\kappa$ in each condition (0.37, 0.51, 0.63, 0.39 for \IQm, \IQp, \QIQp, \QIQm). (C) Vote-free partial correlation between model correctness and each item's mean human accuracy in a condition, controlling for the item's accuracy in the other three. Outlined cells are each model's maximum.}
  \label{fig:supp_e3}
\end{figure*}

The consensus is a majority vote over a median of six participants, so its reliability differs across conditions, and a less reliable reference lowers any agreement computed against it. \Cref{fig:supp_e3} compares raw $\kappa$ with two alternatives.

\paragraph{Reliability rescaling}
Dividing $\kappa$ by $\sqrt{r_c}$ boosts the less reliable brief conditions relative to \QIQp (by a factor of 1.64 for \IQm against 1.26 for \QIQp). Under this rescaling only 4 of 14 autoregressive models keep the \QIQp match. The rescaling is heuristic, because leave-one-out participant agreement is a proxy for, not an estimate of, the reliability of a majority consensus.

\paragraph{Vote-free partial correlation}
This statistic uses each item's mean human accuracy directly and needs no reliability correction. It places 11 of 14 autoregressive models at \QIQp (GPT-6-Luna and \qwenthree at \IQp, nanoLLaVA at \QIQm) and the four contrastive embeddings at a brief condition (\IQm for \siglip and \coca, \QIQm for \clipthirtytwo and \clipsixteen). The two BLIP scorings, which match \IQm under $\kappa$ and $\kappa^{*}$, move to \QIQp.

\paragraph{Reading}
Both alternatives are sensitivity analyses, not bounds on the primary estimate. Together they support a broad contrast between the four contrastive embeddings and the stronger autoregressive models, while showing that individual best matches depend on the estimator.

\FloatBarrier
%% ============================================================================ A6
\section{Controls for the Family Difference}
\label{supp:controls}

This section details the three controls of main \cref{subsec:not_accuracy} and two further checks of the human reference and of model lineage.

\subsection{Difficulty-stratified null}
\label{supp:null}

\begin{figure*}[!ht]
  \centering
  \includegraphics[width=0.95\textwidth]{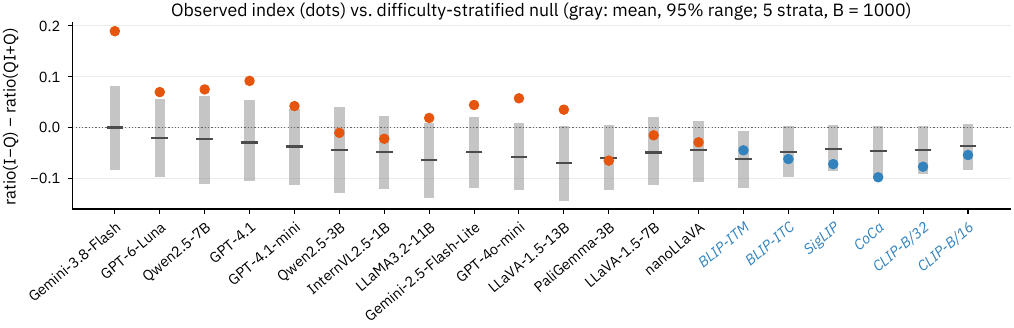}
  \caption{\textbf{Difficulty-stratified permutation null.} Each model's correctness was permuted within five quantile strata of mean human accuracy, preserving its accuracy within every stratum. Gray bars: mean and 95\% range of the index over 1{,}000 permutations; dots: observed index.}
  \label{fig:supp_e19}
\end{figure*}

Permuting a model's correctness within strata of human difficulty keeps its accuracy and the items' shared difficulty, and removes any finer correspondence. Under this null every model has a negative index (\cref{fig:supp_e19}), so zero is not the neutral point. Nine of fourteen autoregressive models, including all the strongest, exceed the null's 95\% range, and one dual encoder (\coca) falls below it. The rank correlation between accuracy and index is 0.88 over all models and 0.77 within the autoregressive family, and the within-family ordering exceeds the null ($p=0.013$).

\subsection{Accuracy-matched item subsets}
\label{supp:matched}

\begin{figure*}[!ht]
  \centering
  \includegraphics[width=0.75\textwidth]{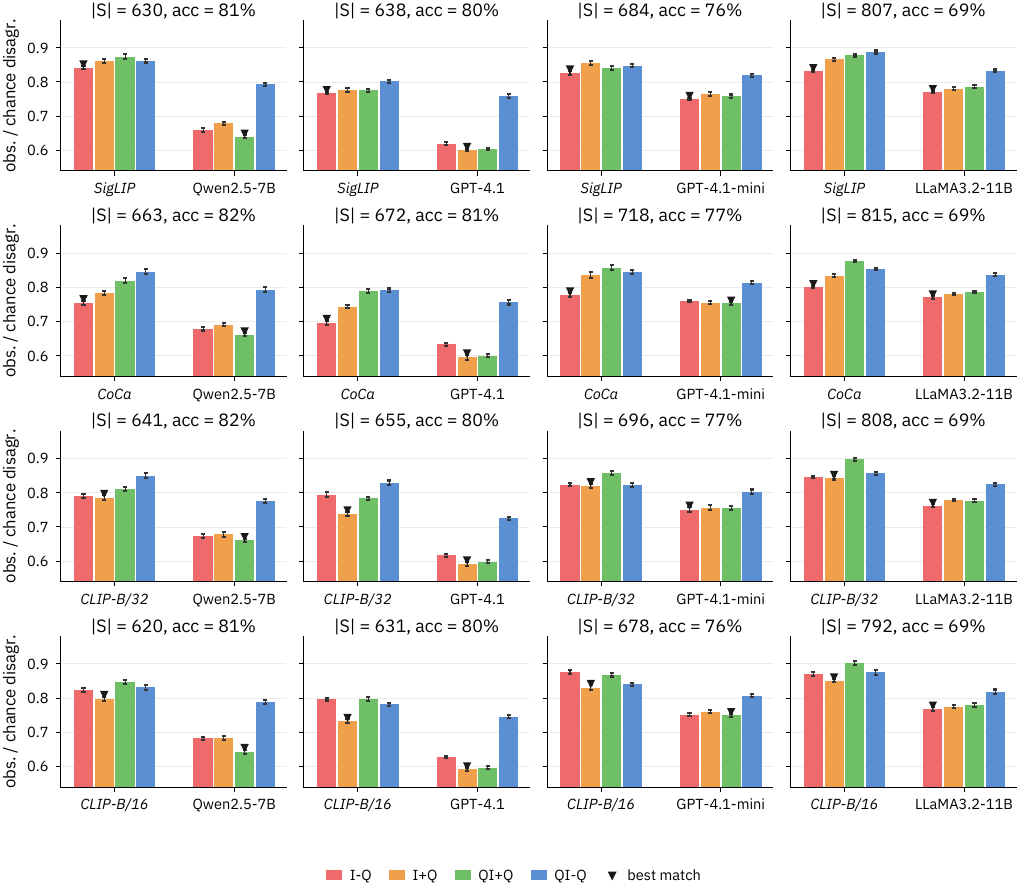}
  \caption{\textbf{Accuracy-matched item subsets.} Each panel pairs one contrastive model (italic) with one strong autoregressive model on identical items with accuracies matched within one point (subset size and matched accuracy above each panel). Bars: SD over 20 removal orders; triangles: best match.}
  \label{fig:supp_e2}
\end{figure*}

\paragraph{Procedure}
We paired each of four strong autoregressive models (\qwenseven, \gptfourone, \gptfourmini, \llamathree) with each of the four contrastive models (\siglip, \coca, \clipthirtytwo, \clipsixteen). For each pair, items on which only the autoregressive model was correct were removed in random order until the accuracies agreed within one point. The resulting subsets hold 620-815 items at 69-83\% accuracy. Twenty random removal orders give the SDs, and chance is re-simulated at the matched accuracy. Because removal depends on model correctness, this is a sensitivity analysis rather than an independent experiment.

\paragraph{Results}
The contrastive model best matches \IQm in 8 of 16 pairs and \IQp in the other 8, never \QIQp (\cref{fig:supp_e2}). The autoregressive model keeps \QIQp in 6 of 16 pairs, including all four with \qwenseven, and moves to \IQp or \IQm in the rest. Both models of a pair show the expected sign of the index in 8 of 16 pairs.

\paragraph{Reverse direction}
Restricting items so that \llavaseven or \gemma reaches the accuracy of \qwenseven (88.3\%) moves \llavaseven to \QIQp (19 of 20 removal orders) and \gemma to \IQp (16 of 20). \llavathirteen, which already matches \QIQp on all items, keeps it in 20 of 20 orders.

\subsection{Answer priors without the image}
\label{supp:blind}

\begin{table*}[!ht]
\centering
\caption{\textbf{Language-prior control, per model.} Withheld: text-only prompt (text) or uniform gray image (gray). Blind index: the index of the image-withheld run. Missed: items with blind correctness below 0.5. Index all / missed: the sighted model's index on all items and on the missed items, with the best match on the missed items. Null: 95\% range of the index under permutations within strata of human difficulty $\times$ blind outcome. Dual encoders in italics.}
\label{tab:supp_blind}
\small
\setlength{\tabcolsep}{4pt}
\begin{tabular}{@{}llrrrrrrlc@{}}
\toprule
Model & Withheld & Acc.\ with & Acc.\ without & Blind index & Missed & Index all & Index missed & Match missed & Null range \\
\midrule
Gemini-3.8-Flash & text & 93.5 & 55.2 & +0.009 & 319 & +0.189 & +0.055 & \IQp & $[-0.07,+0.10]$ \\
GPT-6-Luna & text & 88.4 & 54.3 & -0.003 & 378 & +0.070 & +0.036 & \IQp & $[-0.08,+0.07]$ \\
Qwen2.5-7B & text & 88.3 & 54.2 & -0.024 & 435 & +0.075 & +0.090 & \QIQp & $[-0.11,+0.07]$ \\
GPT-4.1 & text & 87.0 & 59.6 & +0.041 & 358 & +0.092 & +0.071 & \QIQp & $[-0.08,+0.08]$ \\
GPT-4.1-mini & text & 83.5 & 56.0 & +0.016 & 394 & +0.042 & +0.004 & \QIQp & $[-0.10,+0.06]$ \\
Qwen2.5-3B & text & 83.4 & 56.0 & -0.009 & 425 & -0.010 & +0.082 & \QIQp & $[-0.13,+0.04]$ \\
InternVL2.5-1B & text & 75.9 & 52.6 & +0.019 & 453 & -0.022 & -0.048 & \IQp & $[-0.12,+0.03]$ \\
LLaMA3.2-11B & text & 74.9 & 51.3 & -0.022 & 469 & +0.019 & +0.014 & \QIQp & $[-0.13,+0.01]$ \\
Gemini-2.5-Flash-Lite & text & 72.8 & 55.4 & +0.008 & 317 & +0.044 & +0.048 & \IQp & $[-0.10,+0.02]$ \\
GPT-4o-mini & text & 72.4 & 54.6 & -0.005 & 421 & +0.058 & +0.016 & \IQp & $[-0.11,+0.01]$ \\
LLaVA-1.5-13B & text & 72.1 & 54.6 & -0.001 & 432 & +0.035 & +0.040 & \QIQp & $[-0.12,+0.01]$ \\
PaliGemma-3B & gray & 70.0 & 53.9 & -0.009 & 444 & -0.065 & -0.141 & \IQm & $[-0.12,+0.01]$ \\
LLaVA-1.5-7B & text & 68.9 & 51.4 & -0.016 & 463 & -0.015 & +0.026 & \IQp & $[-0.11,+0.01]$ \\
nanoLLaVA & gray & 66.4 & 53.7 & +0.025 & 438 & -0.029 & -0.072 & \QIQm & $[-0.08,+0.03]$ \\
\textit{BLIP-ITM} & gray & 64.2 & 54.2 & +0.046 & 433 & -0.045 & -0.072 & \IQm & $[-0.11,-0.01]$ \\
\textit{BLIP-ITC} & gray & 60.4 & 52.4 & -0.003 & 453 & -0.062 & -0.026 & \IQm & $[-0.10,+0.01]$ \\
\textit{SigLIP} & gray & 59.5 & 49.4 & +0.017 & 492 & -0.072 & -0.072 & \IQm & $[-0.09,+0.01]$ \\
\textit{CoCa} & gray & 59.0 & 51.7 & +0.019 & 463 & -0.098 & -0.061 & \IQm & $[-0.09,+0.01]$ \\
\textit{CLIP-B/32} & gray & 58.6 & 49.6 & +0.013 & 476 & -0.077 & -0.053 & \IQm & $[-0.09,+0.00]$ \\
\textit{CLIP-B/16} & gray & 57.3 & 48.6 & -0.006 & 498 & -0.054 & -0.057 & \QIQm & $[-0.08,+0.01]$ \\
\bottomrule
\end{tabular}
\end{table*}

\paragraph{Procedure}
Every model answered the 960 items a second time with the image withheld. Models that accept text alone (Qwen2.5-7B/3B, LLaMA3.2-11B, LLaVA-1.5-13B/7B, InternVL2.5-1B, the four GPT models and the two Gemini models) received only the question and options. The six dual encoders, PaliGemma-3B and nanoLLaVA require an image and received a uniform mid-gray image at the original size. Templates, decoding and scoring follow the main evaluation (\cref{supp:models}). The Gemini blind run used the Gemini~3.5 Flash-Lite endpoint, as the 2.5 checkpoint was no longer available to our API key, and is compared with the original sighted outputs.

\paragraph{Results}
Without the image, accuracy falls to 49-60\% (57-94\% with it; main \cref{fig:blind}A). The blind runs resemble no condition in particular: their indices lie in $[-0.024, +0.046]$ and their best matches are spread over all four conditions. We then recompute each sighted model's index on the items it misses without the image (317-498 items per model; \cref{tab:supp_blind}). The family pattern is unchanged (main \cref{fig:blind}B): every strong autoregressive model still matches a self-paced condition (six \QIQp, five \IQp), every dual encoder still matches a brief condition, and the family-mean index moves from $+0.054$ to $+0.037$ (autoregressive) and from $-0.068$ to $-0.057$ (dual encoders).

\paragraph{Null that also preserves the language prior}
Permuting sighted correctness within strata of human difficulty crossed with blind outcome (wrong, tie, right; 1{,}000 permutations) preserves accuracy, item difficulty and the language prior together. Seven of fourteen autoregressive models exceed this null's 95\% range (Gemini-3.8-Flash, Qwen2.5-7B, GPT-4.1, LLaMA3.2-11B, Gemini-2.5-Flash-Lite, GPT-4o-mini, LLaVA-1.5-13B), and one dual encoder, \coca, falls below it.

\subsection{Uncertainty in the human reference}
\label{supp:partboot}

\begin{figure*}[!ht]
  \centering
  \includegraphics[width=0.95\textwidth]{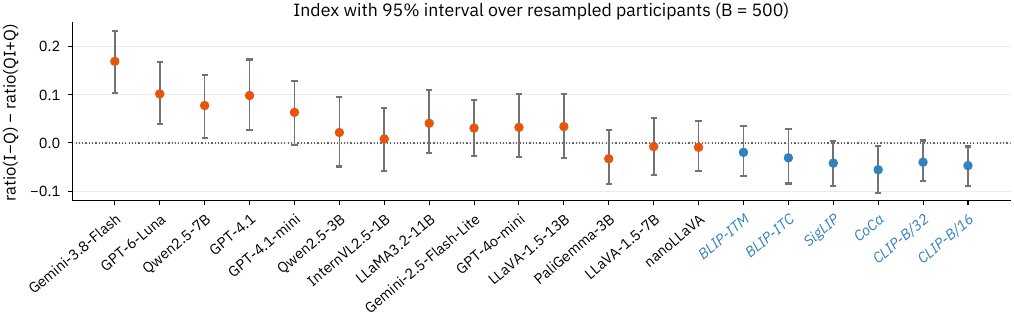}
  \caption{\textbf{Participant bootstrap.} The index with its 95\% interval over 500 resamples of the 100 online participants, with the consensus rebuilt in every draw. Points are bootstrap means, which differ slightly from the fixed-consensus values in \cref{tab:supp_boot}.}
  \label{fig:supp_e20}
\end{figure*}

\Cref{fig:supp_e20} resamples participants rather than fixing the majority-vote labels. Over 500 resamples, the index intervals exclude zero for \coca and \clipsixteen (below) and for Gemini-3.8-Flash, GPT-6-Luna, \qwenseven and \gptfourone (above). No dual encoder has a \QIQp probability above 0.07. The bootstrap median rank correlation between accuracy and index is 0.85 $[0.64, 0.95]$ over all 20 models and 0.77 $[0.49, 0.93]$ within the autoregressive family ($P(\rho>0)=1.000$).

\subsection{Mixed-effects model}
\label{supp:mixed}

\paragraph{Specification}
To account for shared model lineages and benchmark differences, we fit a linear mixed model (statsmodels \texttt{MixedLM}, REML) to the index computed per model and benchmark (96 items each, analytic chance at the model's within-benchmark accuracy). This gives 20 models $\times$ 10 benchmarks, minus three cells (GPT-4.1, Qwen2.5-7B and Gemini-3.8-Flash on WhatsUp) where both model and \QIQp consensus are correct on all 96 items and the ratio is $0/0$. Fixed effects are a dual-encoder indicator and, in a second model, each model's mean-centered overall accuracy. Random effects are an intercept per benchmark and a variance component for lineage (GPT: 6-Luna, 4.1, 4.1-mini, 4o-mini; Gemini: 3.8-Flash, 2.5-Flash-Lite; Qwen: 7B, 3B; LLaVA: 7B, 13B; CLIP: B/16, B/32; BLIP: ITC, ITM; the other six models singletons). Because statsmodels nests variance components within the grouping factor, the lineage term is effectively lineage-by-benchmark.

\paragraph{Results}
The dual-encoder effect is $-0.089$ (SE 0.023, $p=10^{-4}$). With accuracy as a covariate it is $-0.062$ (SE 0.032, $p=0.048$) and the accuracy slope is $+0.16$ (SE 0.13, $p=0.22$). Within the autoregressive family alone ($n=137$) the accuracy slope is $+0.17$ (SE 0.13, $p=0.20$). Variance components are 0.0017 (benchmark), 0.0074 (lineage-by-benchmark) and 0.0098 (residual).

\paragraph{Alternatives}
On the original 18 models, a fully crossed specification gives a dual-encoder effect of $-0.065$ (SE 0.033, $p=0.046$), with the lineage variance at zero. Using within-benchmark instead of overall accuracy as the covariate gives $-0.072$ (SE 0.027, $p=0.008$). Because the families do not overlap in accuracy, the covariate is identified only by within-family spread. The adjusted estimate is therefore a sensitivity check, and the matched-item analysis (\cref{supp:matched}) remains the primary control.

\FloatBarrier
%% ============================================================================ A7
\section{Further Probes}
\label{supp:probes}

This section reports the follow-up analyses summarized in main \cref{subsec:probes}.

\subsection{Answer format}
\label{supp:fullopt}

\begin{figure*}[!ht]
  \centering
  \includegraphics[width=0.8\textwidth]{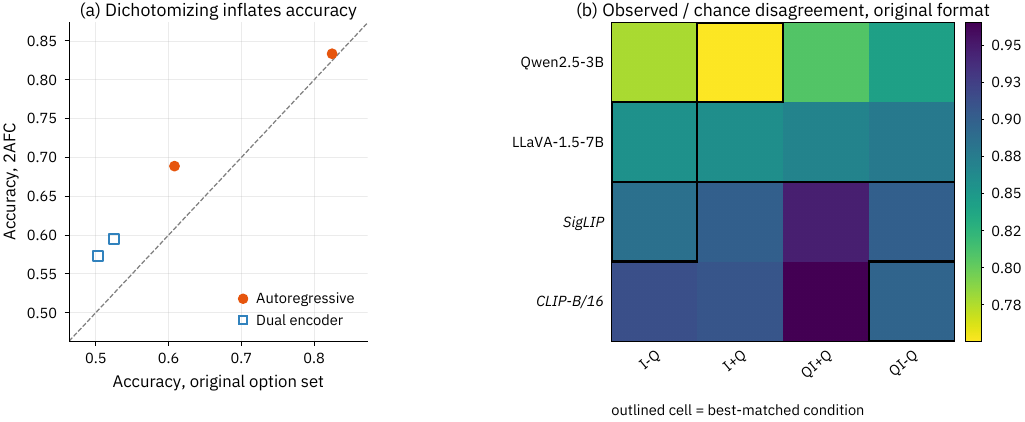}
  \caption{\textbf{Restoring original options on 273 items (models only).} (a) Accuracy with two options against accuracy with the original options. (b) Disagreement ratio per condition with the original options; the outlined best match equals the two-way result for every model.}
  \label{fig:supp_e1}
\end{figure*}

\paragraph{Original options for three benchmarks (273 items)}
Four of the ten benchmarks have more than two native options (\cref{tab:supp_sources}). This model-only rerun restores them for three of these (V*, RealWorldQA and WhatsUp); MME-RealWorld's five-way items were dichotomized when the stimuli were built, and its original options are tested in the full-option human study instead (\cref{supp:fullopt_protocol}). Of the 288 items from these three benchmarks, 273 natively have more than two options (3.7 on average); the other 15 are natively two-way and unchanged. We reran two contrastive and two autoregressive models on all 960 items with the original options restored for those 273, and recomputed each best match against the same two-way human consensus (\cref{fig:supp_e1}, \cref{tab:format}). Accuracy drops by 5.7 points on average, but the best match is unchanged for all four models.

\begin{table}[!ht]
  \centering
  \caption{\textbf{Two-way vs.\ original options on 273 items.} Accuracy and best match on the 960 items when the 273 items with more than two source options are scored in their original format. Contrastive models in italics.}
  \label{tab:format}
  \small
  \begin{tabular}{@{}lcccc@{}}
    \toprule
    & \multicolumn{2}{c}{Accuracy} & \multicolumn{2}{c}{Best match} \\
    \cmidrule(lr){2-3}\cmidrule(lr){4-5}
    Model & 2-way & original & 2-way & original \\
    \midrule
    \qwenthree & 83.4 & 82.5 & \IQp & \IQp \\
    \llavaseven & 68.9 & 60.8 & \IQm & \IQm \\
    \textit{\siglip} & 59.5 & 52.5 & \IQm & \IQm \\
    \textit{\clipsixteen} & 57.3 & 50.3 & \QIQm & \QIQm \\
    \bottomrule
  \end{tabular}
\end{table}

\paragraph{Full-option replication with new human data (192 items)}
The stronger test re-collects the \emph{human} reference in the original format (\cref{supp:fullopt_protocol}). Every model answered the 192 items with their original options under the main templates; the letter hint lists the letters offered, and dual encoders score every option. The two-way answers are the main-evaluation outputs on the same items. For each format, ratios, best match and index are computed against the human reference collected in that format, with chance at the model's accuracy on the 192 items.

\begin{figure*}[!ht]
  \centering
  \includegraphics[width=0.96\textwidth]{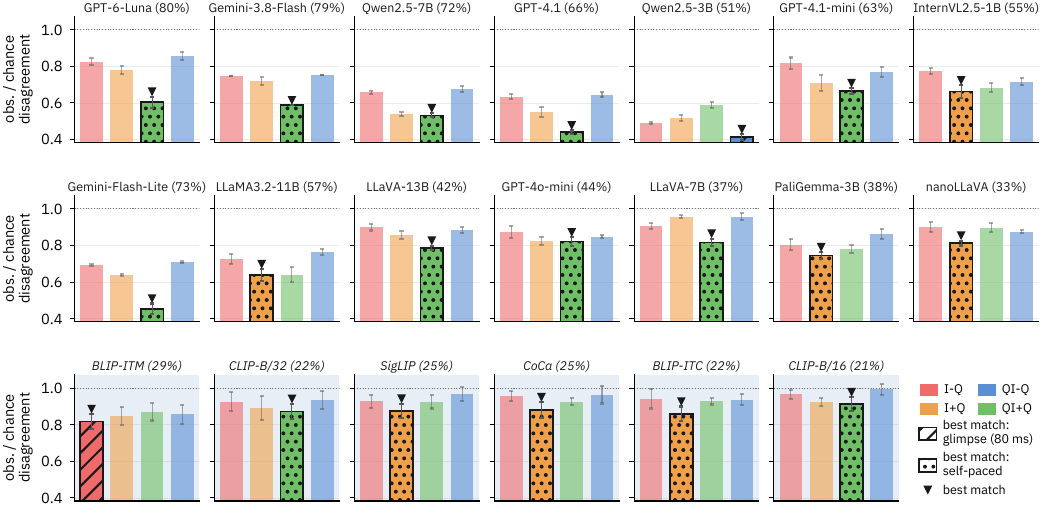}
  \caption{\textbf{Error profiles with the original options.} Disagreement ratio of each model's full-option answers with each human condition of the full-option study, on the 192 items (layout as main \cref{fig:families}).}
  \label{fig:supp_fullopt}
\end{figure*}

\begin{table}[!ht]
\centering
\caption{\textbf{Full-option replication, per model.} Accuracy, best match and index on the 192 items with two options (against the main-study consensus) and with the original options (against the full-option consensus). Dual encoders in italics.}
\label{tab:supp_fullopt}
\small
\begin{tabular}{@{}lrrllrr@{}}
\toprule
 & \multicolumn{2}{c}{Accuracy (\%)} & \multicolumn{2}{c}{Best match} & \multicolumn{2}{c}{Index} \\
\cmidrule(lr){2-3}\cmidrule(lr){4-5}\cmidrule(lr){6-7}
Model & two & original & two & original & two & original \\
\midrule
GPT-6-Luna & 89.7 & 80.3 & \QIQp & \QIQp & +0.15 & +0.22 \\
Gemini-3.8-Flash & 89.4 & 79.2 & \QIQp & \QIQp & +0.13 & +0.16 \\
Qwen2.5-7B & 86.4 & 72.1 & \QIQp & \QIQp & +0.15 & +0.12 \\
GPT-4.1 & 83.5 & 66.3 & \QIQp & \QIQp & +0.15 & +0.19 \\
Qwen2.5-3B & 81.7 & 51.3 & \QIQp & \QIQm & +0.12 & -0.10 \\
GPT-4.1-mini & 81.4 & 62.8 & \QIQp & \QIQp & +0.23 & +0.15 \\
InternVL2.5-1B & 75.3 & 54.6 & \IQp & \IQp & +0.13 & +0.09 \\
Gemini-2.5-Flash-Lite & 73.8 & 72.7 & \QIQp & \QIQp & +0.27 & +0.24 \\
LLaMA3.2-11B & 72.2 & 56.6 & \QIQp & \IQp & +0.26 & +0.09 \\
LLaVA-1.5-13B & 70.0 & 42.5 & \QIQp & \QIQp & +0.13 & +0.11 \\
GPT-4o-mini & 67.3 & 44.1 & \QIQp & \QIQp & +0.18 & +0.05 \\
LLaVA-1.5-7B & 66.0 & 37.2 & \IQp & \QIQp & -0.00 & +0.09 \\
PaliGemma-3B & 59.7 & 37.8 & \QIQp & \IQp & +0.12 & +0.02 \\
nanoLLaVA & 52.1 & 33.3 & \QIQm & \IQp & +0.01 & +0.01 \\
\textit{BLIP-ITM} & 54.0 & 28.7 & \IQm & \IQm & -0.06 & -0.05 \\
\textit{CLIP-B/32} & 53.8 & 22.4 & \QIQm & \QIQp & -0.09 & +0.05 \\
\textit{SigLIP} & 50.4 & 25.4 & \IQm & \IQp & -0.05 & +0.00 \\
\textit{CoCa} & 50.0 & 25.2 & \IQm & \IQp & -0.10 & +0.03 \\
\textit{BLIP-ITC} & 49.2 & 21.9 & \IQp & \IQp & +0.06 & +0.01 \\
\textit{CLIP-B/16} & 46.5 & 21.1 & \QIQm & \QIQp & -0.04 & +0.05 \\
\bottomrule
\end{tabular}
\end{table}

\paragraph{Human results}
Human accuracy on these items is 41.1\% (\IQm), 41.5\% (\QIQm), 51.0\% (\IQp) and 57.8\% (\QIQp) with the original options, against 59.2, 62.3, 67.2 and 73.6\% with two options (main-study participants, same items). Condition contrasts (main \cref{fig:format}B) use each condition's per-item accuracy and a paired item bootstrap; values are two-way, then original:
\QIQp $-$ \IQm $+14.4$ [$+10.4$, $+18.5$] and $+16.7$ [$+13.1$, $+20.3$];
\IQp $-$ \IQm $+8.0$ [$+3.7$, $+12.0$] and $+9.9$ [$+6.6$, $+13.1$];
\QIQm $-$ \IQm $+3.1$ [$-0.5$, $+6.8$] and $+0.4$ [$-2.5$, $+3.3$];
\QIQp $-$ \QIQm $+11.3$ [$+7.3$, $+15.6$] and $+16.3$ [$+12.5$, $+20.3$];
\QIQp $-$ \IQp $+6.4$ [$+2.8$, $+10.0$] and $+6.8$ [$+3.8$, $+9.9$].
The between-format differences are $+2.3$ [$-2.1$, $+7.0$], $+1.9$ [$-3.2$, $+7.0$], $-2.7$ [$-7.3$, $+1.9$], $+5.0$ [$+0.3$, $+9.8$] and $+0.4$ [$-4.1$, $+4.8$]; only the \QIQp $-$ \QIQm difference excludes zero.

\paragraph{Model results}
Mean model accuracy falls from 74.9 to 56.5\% for the autoregressive models and from 50.6 to 24.1\% for the dual encoders (chance 25\%). The autoregressive mean index is $+0.14$ [$+0.06$, $+0.23$] with two options and $+0.11$ [$+0.05$, $+0.17$] with the original options; the dual-encoder mean is $-0.05$ [$-0.11$, $+0.02$] and $+0.02$ [$-0.04$, $+0.08$]. The family gap shrinks from 0.19 [0.08, 0.30] to 0.09 [0.02, 0.17], because the dual encoders, now at chance, no longer carry a profile. The change of $-0.10$ has an item-bootstrap interval of [$-0.21$, $+0.02$], and [$-0.17$, $-0.03$] when resampling models within families. Across the 20 models, the index is rank-correlated between formats at 0.68 [0.04, 0.76] (Pearson 0.56 [0.03, 0.75]).

\paragraph{Which best matches change}
Thirteen of fourteen autoregressive models keep a positive index and thirteen a self-paced best match. The best match changes for 9 of 20 models. Seven of these are four dual encoders and the three weakest autoregressive models, whose four ratios all lie within 0.25 of 1 in both formats, so their best match is close to arbitrary (the gap to the runner-up is 0.00-0.07). The other two are \llamathree, which moves between the two self-paced conditions with a runner-up gap below 0.01, and \qwenthree, whose index changes sign ($+0.12$ to $-0.10$). For the other autoregressive models the smallest ratio is 0.44-0.82, and where the runner-up lies within 0.02 it is the other self-paced condition.

\paragraph{Bootstrap details}
The item bootstrap (2{,}000 draws) resamples the 192 items once per draw for all models and both formats. Its chance term is the closed-form expectation of the simulation of \cref{supp:aggregation}, and the resulting ratios agree with \cref{tab:supp_fullopt} to within 0.01.

\subsection{Image resolution, shared encoders and question-guided crops}
\label{supp:resolution}

\begin{figure}[!ht]
  \centering
  \includegraphics[width=0.8\textwidth]{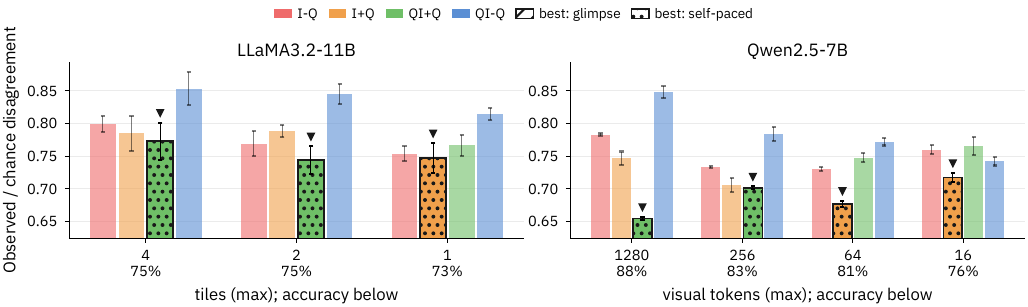}
  \caption{\textbf{Resolution sweeps.} Disagreement ratio with each human condition as the maximum number of image tiles (\llamathree) or visual tokens (\qwenseven) is reduced; best match outlined, SD across the three sweep templates, accuracy under each setting.}
  \label{fig:supp_resolution}
\end{figure}

\paragraph{Resolution sweeps}
We reduced the maximum number of visual tokens of \qwenseven (1{,}280, 256, 64, 16), the number of $560^2$ image tiles of \llamathree (4, 2, 1) and the tiles of InternVL2.5-1B (12, 4, 1), each under three templates (\cref{fig:supp_resolution}; lower block of \cref{tab:supp_boot}). Because of the different template count, the full-resolution index of \qwenseven ($+0.12$) differs from the five-template main value ($+0.07$). For \qwenseven the index falls from $+0.12$ $[+0.02, +0.22]$ at 1{,}280 tokens (88\% accuracy) to near zero at 64 and 16 tokens (81\% and 76\%), and the best match moves from \QIQp (1{,}280 and 256 tokens) to \IQp (64 and 16). \llamathree matches \QIQp at 4 and 2 tiles (75\%) and \IQp at one tile (73\%), with the bootstrap probability of \QIQp falling from 0.54 and 0.70 to 0.14. The shift is toward free viewing, not a glimpse. It does not separate the loss of detail from the accompanying drop in accuracy.

\begin{table}[!ht]
\centering
\caption{\textbf{Same vision encoder, different readout.} Accuracy, index with 95\% item-bootstrap interval, and best match for CLIP ViT-L/14-336 scored contrastively and for \llavaseven and \llavathirteen, which read the same encoder's patch tokens with a language model; and likewise for SigLIP, \gemma and nanoLLaVA. Dual encoders in italics.}
\label{tab:backbone}
\small
\setlength{\tabcolsep}{5pt}
\begin{tabular}{@{}lrrl@{}}
\toprule
Model & Acc.\ (\%) & Index [95\% CI] & Best \\
\midrule
\multicolumn{4}{@{}l}{\emph{Encoder: CLIP ViT-L/14-336}} \\
\textit{CLIP (contrastive)} & 60.0 & $-0.06$ $[-0.11,\,-0.01]$ & \IQm \\
LLaVA-1.5-7B & 68.9 & $-0.02$ $[-0.09,\,+0.06]$ & \IQm \\
LLaVA-1.5-13B & 72.1 & $+0.04$ $[-0.04,\,+0.12]$ & \QIQp \\
\midrule
\multicolumn{4}{@{}l}{\emph{Encoder: SigLIP}} \\
\textit{SigLIP (contrastive)} & 59.5 & $-0.07$ $[-0.12,\,-0.03]$ & \IQm \\
PaliGemma-3B & 70.0 & $-0.07$ $[-0.13,\,+0.01]$ & \IQm \\
nanoLLaVA & 66.4 & $-0.03$ $[-0.09,\,+0.03]$ & \QIQm \\
\bottomrule
\end{tabular}
\end{table}

\paragraph{Same encoder, different readout}
\Cref{tab:backbone} holds the vision encoder fixed. CLIP ViT-L/14-336, the OpenAI checkpoint used inside LLaVA-1.5, scored contrastively through its pooled embedding has an index of $-0.059$ $[-0.11, -0.01]$ and a \QIQp probability of 0.00; the 224-px ViT-L/14 gives $-0.071$ $[-0.12, -0.02]$. The same encoder's patch tokens read by \llavaseven give $-0.015$ $[-0.09, +0.06]$, and by \llavathirteen $+0.035$ $[-0.04, +0.12]$ with \QIQp as best match (probability 0.63). On the SigLIP encoder, the contrastive model gives $-0.072$, nanoLLaVA $-0.029$ and \gemma $-0.065$. Overlapping intervals and differences in training make this comparison suggestive rather than a controlled intervention.

\begin{figure*}[!ht]
  \centering
  \includegraphics[width=0.85\textwidth]{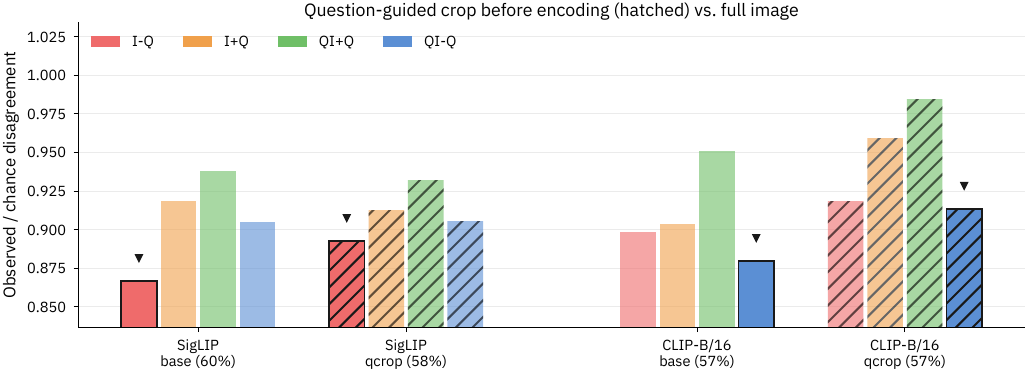}
  \caption{\textbf{A question-guided crop does not move contrastive models.} Disagreement ratio per condition for two contrastive models answering from the full image (plain) or from a question-selected crop (hatched); accuracy in parentheses.}
  \label{fig:supp_e9}
\end{figure*}

\paragraph{Question-guided crop}
To test whether a dual encoder can be moved off the glimpse end by giving it the relevant region before encoding, we scored 14 candidate windows (the full image, nine half-size windows at quarter stride and four at 0.7 scale) against the question text with CLIP ViT-B/16, and gave each model the best-scoring window (\cref{fig:supp_e9}). Accuracy and best match are unchanged for \siglip (\IQm) and \clipsixteen (\QIQm). This negative result applies to these CLIP-selected crops; it does not rule out better localization or other forms of question conditioning.

\subsection{Instructed emulation of a viewing condition}
\label{supp:emulate}

\paragraph{Motivation and procedure}
If the error profile reflected knowledge about viewers rather than the model's own access to the image, a model could be asked to reproduce it. Each model answered the 960 items four more times, with the image shown and the question and options unchanged, but with an instruction to predict what a person in a stated condition chose. For example, for \IQm: ``A human subject saw this image for only 80 milliseconds, followed by a visual-noise mask, and only then read the question and options''; the \IQp version says the person ``viewed this image for as long as they wanted''; the \QIQm version that they ``first read the question, then saw this image for only 80 milliseconds \ldots and then answered''; and \QIQp the self-paced variant. Each is followed by ``Predict which option this subject chose'' and the template's answer-format sentence, with ``the correct answer'' replaced by ``the subject's answer''. Dual encoders, whose text encoders truncate at 64-77 tokens, receive a one-sentence version prepended to every option (\eg ``A person who glimpsed the image for 80 ms before reading the question answered:''). Twelve models have all four runs.

\begin{figure}[!ht]
  \centering
  \includegraphics[width=0.55\textwidth]{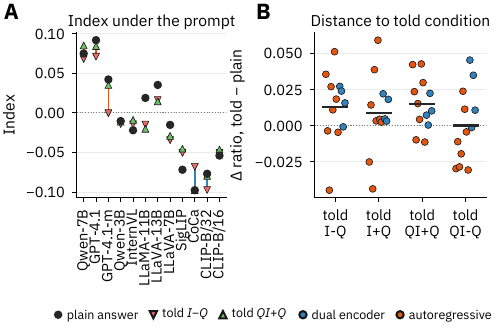}
  \caption{\textbf{Instructed emulation.} (A) Index of each model's plain answers (black) and of its predictions under the \IQm (red) and \QIQp (green) instructions. (B) Change in chance-corrected disagreement with the stated condition's humans, instruction minus plain answer; one dot per model and condition, bars are means.}
  \label{fig:supp_hpred}
\end{figure}

\begin{table}[!ht]
\centering
\caption{\textbf{Told to emulate a viewer, models do not.} Best match of each model's plain answers and of its predictions under each stated condition; $\checkmark$ marks a prediction whose best match is the stated condition. Shift: index under the \QIQp instruction minus under the \IQm instruction. $\Delta$: mean change in chance-corrected disagreement with the stated condition's humans, instruction minus plain (negative = closer). Dual encoders in italics.}
\label{tab:emulate}
\small
\setlength{\tabcolsep}{4pt}
\begin{tabular}{@{}llllllrr@{}}
\toprule
 & & \multicolumn{4}{c}{told\ldots} & & \\
\cmidrule(lr){3-6}
Model & plain & \IQm & \IQp & \QIQm & \QIQp & shift & $\Delta$ \\
\midrule
Qwen2.5-7B & \QIQp & \IQp & \IQp$^{\checkmark}$ & \IQp & \IQp & +0.02 & +0.006 \\
GPT-4.1 & \QIQp & \IQp & \QIQp & \IQp & \IQp & +0.01 & +0.006 \\
GPT-4.1-mini & \QIQp & \IQp & \IQp$^{\checkmark}$ & \IQp & \QIQp$^{\checkmark}$ & +0.04 & -0.032 \\
Qwen2.5-3B & \IQp & \IQp & \IQp$^{\checkmark}$ & \IQp & \IQp & +0.00 & +0.015 \\
InternVL2.5-1B & \IQp & \IQp & \IQp$^{\checkmark}$ & \IQp & \IQp & +0.01 & +0.011 \\
LLaMA3.2-11B & \QIQp & \IQm$^{\checkmark}$ & \IQm & \IQp & \IQm & -0.01 & +0.041 \\
LLaVA-1.5-13B & \QIQp & \IQp & \IQp$^{\checkmark}$ & \IQp & \IQp & -0.00 & +0.012 \\
LLaVA-1.5-7B & \IQm & \IQm$^{\checkmark}$ & \IQm & \IQm & \IQm & +0.01 & -0.011 \\
\textit{SigLIP} & \IQm & \IQm$^{\checkmark}$ & \IQm & \IQp & \IQm & +0.01 & +0.026 \\
\textit{CoCa} & \IQm & \QIQm & \IQm & \IQm & \IQm & -0.03 & +0.007 \\
\textit{CLIP-B/32} & \IQm & \IQm$^{\checkmark}$ & \IQm & \IQm & \IQm & +0.02 & +0.015 \\
\textit{CLIP-B/16} & \QIQm & \QIQm & \QIQm & \QIQm$^{\checkmark}$ & \QIQm & +0.00 & +0.014 \\
\bottomrule
\end{tabular}
\end{table}

\paragraph{Measures}
For each run we compute the best match and index against the four consensuses, the fraction of items whose template-averaged answer differs from the plain answer, the accuracy of the predicted answers, and the chance-corrected disagreement with the stated condition's humans compared with that of the plain answer (500-draw item bootstrap of the difference).

\paragraph{Results}
The instruction changes 9-60\% of answers, but not the profile (\cref{tab:emulate}, \cref{fig:supp_hpred}). The index moves by less than 0.04 between the \IQm and \QIQp instructions for every model. Most autoregressive models match \IQp whatever condition they are told to emulate. Predicted answers are on average \emph{further} from the stated condition's humans than the plain answers; no model-condition pair moves closer with an interval excluding zero (the closest, \gptfourmini told \IQm or \IQp, have upper bounds at zero). Finally, the accuracy of the predicted answers varies by at most 3 points across stated conditions, whereas human accuracy varies by 14.

\subsection{Per-benchmark co-errors}
\label{supp:perbench}

\begin{figure*}[!ht]
  \centering
  \includegraphics[width=0.95\textwidth]{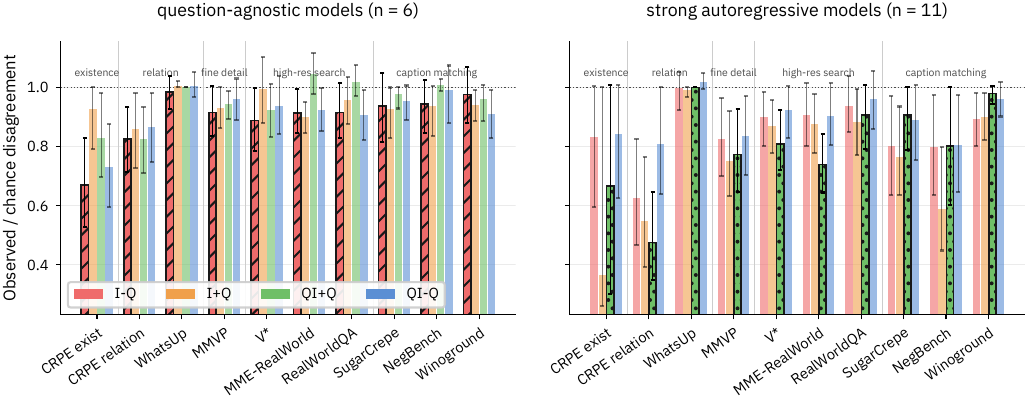}
  \caption{\textbf{Per-benchmark disagreement ratios.} Family means for the six question-agnostic models (left) and the eleven strong autoregressive models (right), with chance at each model's within-benchmark accuracy; bars are 95\% item-bootstrap intervals; each family's signature condition is saturated.}
  \label{fig:supp_e13}
\end{figure*}

\Cref{fig:supp_e13} breaks the family profiles down by benchmark. Question-agnostic models agree with \IQm most on CRPE existence and relation, less on the high-resolution search benchmarks, and little on SugarCrepe, NegBench, Winoground or WhatsUp. Strong autoregressive models agree with \QIQp most on CRPE relation, MME-RealWorld, V* and MMVP. The family pattern therefore does not extend uniformly to every benchmark. WhatsUp is grouped with CRPE relation as a relation benchmark, since its captions differ only in the spatial relation between two objects. The per-benchmark human gap between \QIQp and \IQm does not predict where shared errors concentrate (rank correlation $\approx0$ for both families over the ten benchmarks), so the pattern is not a by-product of where the condition mattered most to people.

\FloatBarrier
%% ============================================================================ A8
\section{Attribution and Gaze}
\label{supp:gaze}

This section reports the preliminary gaze comparison mentioned in main \cref{subsec:probes}. We compare D-RISE attribution maps~\cite{petsiuk2021black} of two dual-encoder (\clipsixteen, \siglip) and two autoregressive models (\qwenthree, \llamathree) with webcam gaze maps of in-person participants under \IQp and \QIQp. \IQm gaze would be a single fixation and is omitted.

\begin{figure*}[!ht]
  \centering
  \includegraphics[width=0.98\textwidth]{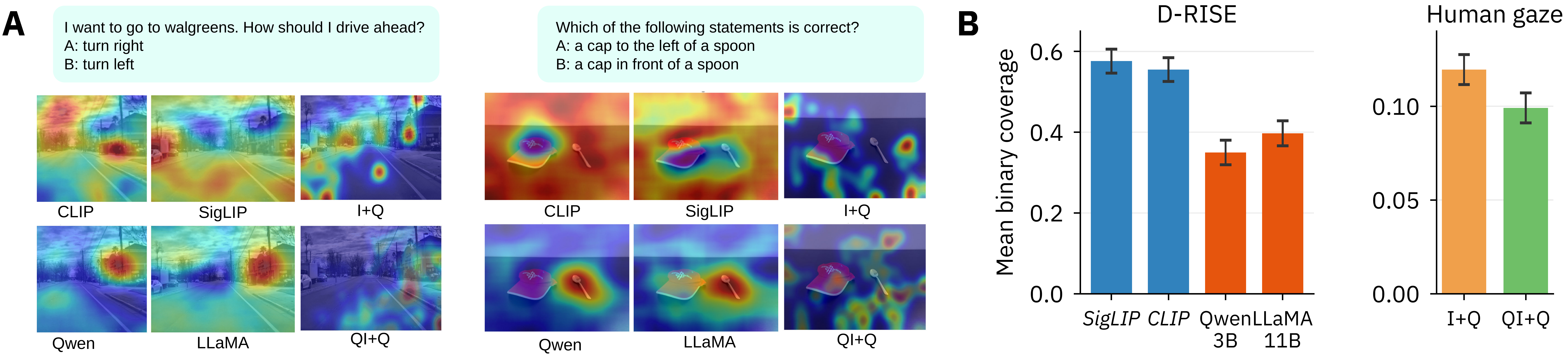}
  \caption{\textbf{Model attribution next to human gaze.} (A) D-RISE attribution for two dual encoders (top row of each grid) and two autoregressive models (bottom row), next to gaze maps averaged over 3-5 in-person participants (\IQp above, \QIQp below), for a RealWorldQA item (left; answer A) and a WhatsUp item (right; answer B). Red marks regions that hurt the model when masked, or that attracted gaze. (B) Mean binary coverage of attribution and gaze maps (95\% CIs; scales differ between map types).}
  \label{fig:supp_gaze_main}
\end{figure*}

\paragraph{Summary}
In the examples, dual-encoder attribution and image-first gaze are diffuse, whereas autoregressive attribution and question-first gaze concentrate on question-relevant regions (\cref{fig:supp_gaze_main}A). Binary coverage, the fraction of the image in the top half of a map's range, quantifies this spread (\cref{fig:supp_gaze_main}B). The comparison concerns spread within each map type, not spatial overlap between maps, and we read it as preliminary, qualitative corroboration rather than independent validation.

\paragraph{Human gaze maps}
Each trial provides gaze coordinates in screen space and the bounding box of the image. Samples were converted to image pixel coordinates (with a vertical flip for Gorilla's bottom-left origin), and samples outside the image were discarded. Per participant and image, samples were rasterized into a count map and smoothed with a Gaussian of $\sigma=0.03\sqrt{W^2+H^2}$ to absorb webcam noise. Participant maps were max-normalized and averaged over at least three participants per image. The 20 in-person participants completed 4{,}800 gaze trials. The calibration routine reports pass or fail against the $4^\circ$ criterion rather than a numeric error (five participants needed a second calibration), so per-participant error is unavailable. The 90 items used here are those viewed by at least three in-person participants in each of \IQp and \QIQp (4-12 participants per item in \IQp, mean 7.0; 3-11 in \QIQp, mean 8.7).

\paragraph{D-RISE}
For each item we generated $N$ random binary masks $M_i$ on a $7\times7$ grid with keep probability 0.5, bilinearly upsampled to full resolution and multiplied into the image (seed 0; no random translation). For each masked input we recorded a score $f_i$ for the correct answer under template~1. For question-agnostic models this is the cosine similarity between the masked-image embedding and the text embedding of the question with the correct option (BLIP-ITM: its matching probability). For autoregressive models it is the logit of the first token of the correct option letter at the answer position. Scores are min-max normalized over masks, the map is $S(x)=\sum_i f_iM_i(x)/\sum_iM_i(x)$, min-max normalized to $[0,1]$, and at least eight finite scores are required. \Cref{fig:supp_gaze_main} uses $N=2{,}000$ masks. Maps can optionally be thresholded at $\tau\in\{0,25,40,50,60,75\}$ percent of their range for display.

\paragraph{Binary coverage}
For a map $A(x)$ over pixels $x\in\Omega$, non-finite values are set to 0 and the map is min-max normalized to $\tilde{A}$ (a constant map is set to zero). Coverage at threshold $\tau$ is $\mathrm{BC}_\tau(A)=\frac{1}{|\Omega|}\sum_{x\in\Omega}\mathbf{1}[\tilde{A}(x)\ge\tau]$; we use $\tau=0.5$. Model coverage is averaged over the items shown to in-person participants in \IQp and in \QIQp, and the two means are combined weighted by the number of items. Human bars average per-image coverage over participants. Absolute values depend on smoothing and threshold, so coverage is comparable within, not across, map types.

\begin{figure}[!ht]
  \centering
  \includegraphics[width=0.5\textwidth]{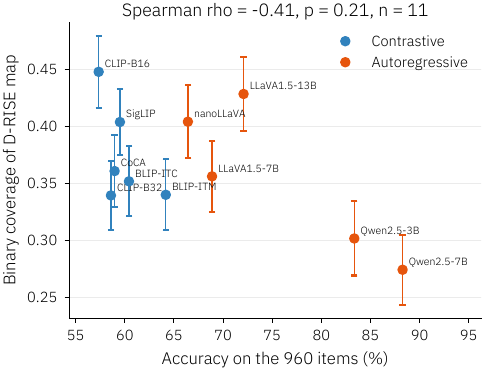}
  \caption{\textbf{Attribution concentration across models.} Mean binary coverage of D-RISE maps (400 masks; SEM over the 90 in-person items) against accuracy for the eleven open-weight models with maps on these items. The API models cannot be masked at this scale; \llamathree, \gemma and InternVL2.5-1B were not run on these items.}
  \label{fig:supp_coverage}
\end{figure}

\paragraph{Coverage across models}
\Cref{fig:supp_coverage} extends coverage to eleven models with lighter maps (400 instead of 2{,}000 masks, hence lower absolute values that should not be compared with \cref{fig:supp_gaze_main}B). \qwenseven and \qwenthree draw on the most concentrated regions (coverage 0.27-0.30). The other nine, dual-encoder and autoregressive alike, span 0.34-0.45 with no ordering, and the rank correlation with accuracy is not significant ($\rho=-0.41$, $n=11$).

\begin{figure*}[!ht]
  \centering
  \includegraphics[width=0.8\textwidth]{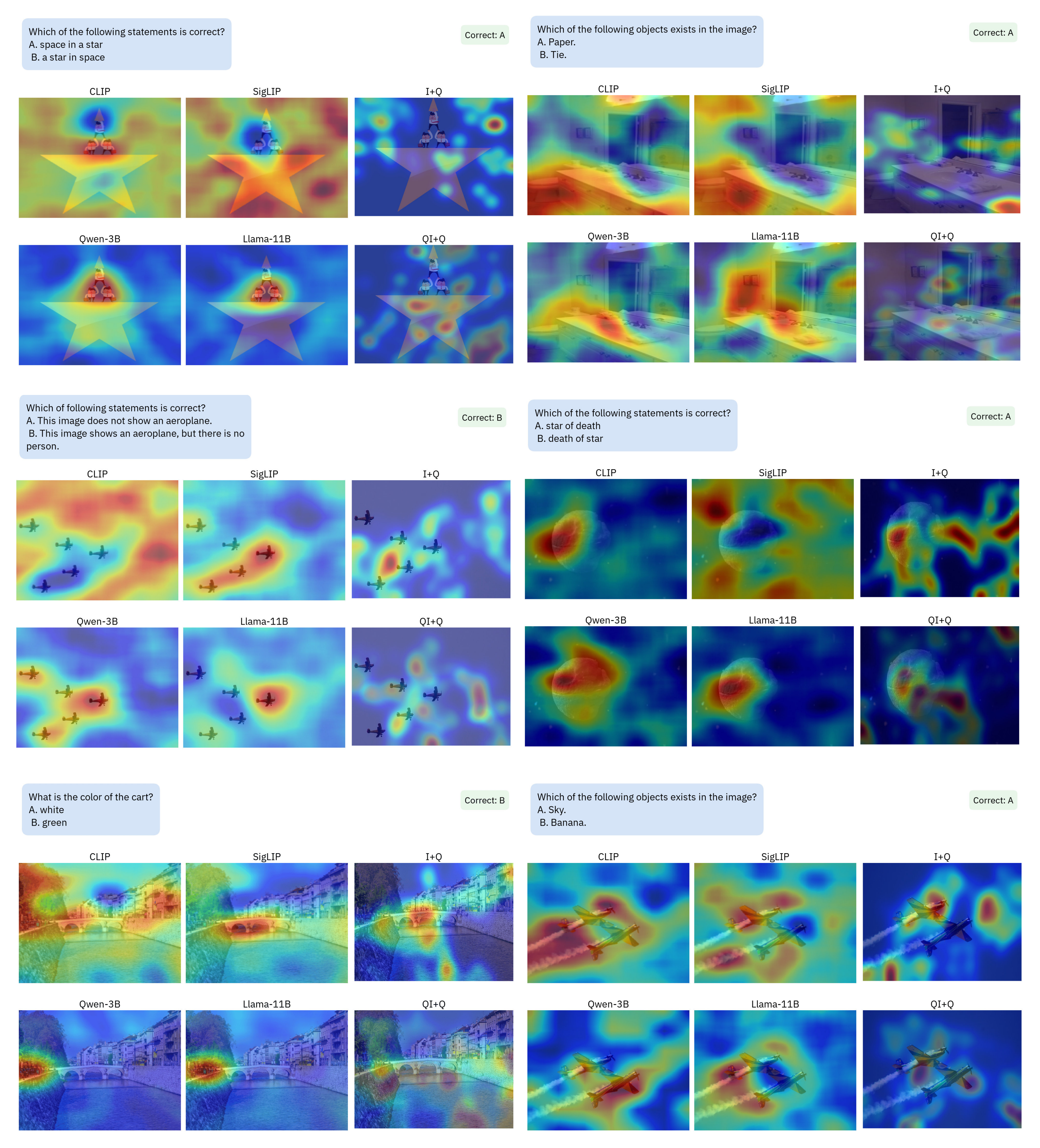}
  \caption{\textbf{Additional gaze-attribution examples.} For each item, the top row shows D-RISE attribution for \clipsixteen and \siglip next to \IQp gaze; the bottom row shows \qwenthree and \llamathree next to \QIQp gaze. Gaze maps average 3-5 participants; red marks high attribution or gaze density.}
  \label{fig:supp_examples}
\end{figure*}

\Cref{fig:supp_examples} shows additional examples.

\FloatBarrier
%% ============================================================================ A9
\section{Datasheet and Release}
\label{sec:supp_datasheet}
We follow the datasheet structure of Gebru et al.~\cite{gebru2021datasheets}.

\paragraph{Motivation} \dsname supplies human VQA baselines under stated, reproducible viewing conditions, and allows item-level comparison of model and human errors against more than one human reference. It was collected by the authors; no third party funded or directed its design.

\paragraph{Composition}
\begin{itemize}[leftmargin=*,nosep]
  \item \emph{Stimuli:} 960 image-question pairs, 96 sampled at random from each of ten benchmark subsets (\cref{tab:supp_sources}), with the source options and the two-way options used here.
  \item \emph{Human responses:} 28{,}800 two-way trials (24{,}000 online, 4{,}800 in person) and 6{,}336 full-option trials (33 further online participants, 192 items), each with anonymized participant id, setting, condition, exposure, mask, image size, response, correctness and reaction time. Every item is covered in every condition (4-9 online responses per item and condition).
  \item \emph{Auxiliary human data:} NASA-TLX ratings per block, and WebGazer gaze samples at 30~Hz for the 20 in-person participants.
  \item \emph{Model outputs:} the answer and correctness of 20 VLMs on every item under every template, plus the resolution sweeps, the \clipl run, the question-guided crops, the original-option reruns, the image-withheld runs, the full-option runs and the emulation runs.
  \item \emph{Audit:} the table for the 101 benchmark reports.
\end{itemize}
No participant-identifying information, video or raw webcam frames are included.

\paragraph{Collection} Online participants were recruited on Prolific and tested in Gorilla (\cref{supp:protocol}); in-person participants were tested on one calibrated workstation. All studies ran under an IRB-approved protocol with informed consent, and participants were paid above local minimum wage. Online data were collected in February 2026, in-person data in October 2026 and the full-option study in November 2026.

\paragraph{Preprocessing} Trials without an A or B key press are dropped (36 of 24{,}000 online). The per-item consensus is the majority vote over online participants, with ties counted as correct (\cref{supp:consensus}). Model outputs that are not a single option letter are scored incorrect for that template (\cref{supp:aggregation}). Raw Gorilla exports and raw model outputs are released alongside the processed tables.

\paragraph{Uses} Condition-specific human baselines for any benchmark overlapping the item pool; item-level human difficulty and consensus for sample-level model evaluation; model-human error comparison; and a template for collecting comparable baselines. The data must not be used to identify participants or infer individual traits. The 960 items are a sample of their benchmarks and do not replace the source leaderboards.

\paragraph{Distribution and licensing} Human responses, gaze, workload and timing data, model outputs, the audit table, the Gorilla experiment variants and the analysis code will be hosted in a public repository with a DOI upon full release. All released materials are distributed under CC~BY~4.0. Stimuli are redistributed under the terms of their sources (\cref{tab:supp_sources}); items whose images may not be redistributed are released as identifiers with a download script.

\paragraph{Maintenance} The authors will maintain the repository, record corrections in a changelog, and accept new human-condition or model-output contributions that follow the released protocol.

\begin{table}[!ht]
  \centering
  \footnotesize
  \setlength{\tabcolsep}{4pt}
  \begin{tabular}{@{}llcl@{}}
    \toprule
    Source & Items & Options & Source license / terms as listed \\
    \midrule
    MMVP~\cite{tong_eyes_2024} & 96 & 2 & MIT (repository and dataset card); images from ImageNet and LAION \\
    MME-RealWorld-Lite~\cite{zhang_mme-realworld_2025} & 96 & 5 & Apache-2.0 (dataset card) \\
    WhatsUp~\cite{kamath_whats_2023} & 96 & 4 & MIT (repository) \\
    NegBench~\cite{alhamoud_vision-language_2025} & 96 & 2 & MIT (repository); images from COCO and VOC \\
    SugarCrepe~\cite{hsieh_sugarcrepe_2023} & 96 & 2 & MIT; images from COCO \\
    Winoground~\cite{thrush_winoground_2022} & 96 & 2 & gated, research only; Getty Images \\
    RealWorldQA~\cite{realworldqa2024} & 96 & 2-4 & CC~BY-ND~4.0 (dataset card and release note) \\
    CRPE existence~\cite{wang2024allseeingprojectv2generalCRPE} & 96 & 2 & Apache-2.0 (dataset card) \\
    CRPE relation~\cite{wang2024allseeingprojectv2generalCRPE} & 96 & 2 & Apache-2.0 (dataset card) \\
    V*~\cite{wu_v_2023} & 96 & 2-4 & MIT; images from SA-1B \\
    \bottomrule
  \end{tabular}
  \caption{\textbf{Stimulus sources.} Native number of options per item and the license under which each source distributes its data. Winoground images are released by identifier only, and RealWorldQA items, whose license forbids modified copies, as identifiers with our two-way option labels.}
  \label{tab:supp_sources}
\end{table}

\clearpage
\end{document}